\documentclass[5p,twocolumn]{elsarticle}
\makeatletter
\def\ps@pprintTitle{%
	\let\@oddhead\@empty
	\let\@evenhead\@empty
	\def\@oddfoot{\rightline{\thepage}}%
	\let\@evenfoot\@oddfoot}
\makeatother

\usepackage{lineno,hyperref}
\usepackage[justification=centering]{caption} 
\usepackage[export]{adjustbox}
\usepackage{subcaption}
\usepackage{amsmath}
\usepackage{amsfonts}
\usepackage[boxed]{algorithm2e}
\usepackage{booktabs}
\usepackage{multirow}
\usepackage{color,soul}  % Text highlight
\usepackage{mathtools, nccmath}
\usepackage{amssymb}
\usepackage{pifont}
\usepackage{hyperref}

\soulregister\cite7      % For citation highlight
\soulregister\ref7		 % For reference highlight
\modulolinenumbers[5]

\journal{Journal of Applied Soft Computing}

\begin{document}

\begin{frontmatter}

\title{Safety-enhanced UAV Path Planning with Spherical Vector-based Particle Swarm Optimization}

\author[1,2]{Manh Duong Phung\corref{cor1}}
\ead{manhduong.phung@uts.edu.au}

\author[1]{Quang Phuc Ha}
\ead{quang.ha@uts.edu.au}

\address[1]{School of Electrical
	and Data Engineering, University of Technology Sydney (UTS) \\
	15 Broadway, Ultimo NSW 2007, Australia}
\address[2]{VNU University of Engineering and Technology (VNU-UET), Vietnam National University, Hanoi (VNU) \\ 144 Xuan Thuy, Cau Giay, Hanoi, Vietnam}

\cortext[cor1]{Corresponding author}

\begin{abstract}
This paper presents a new algorithm named spherical vector-based particle swarm optimization (SPSO) to deal with the problem of path planning for unmanned aerial vehicles (UAVs) in complicated environments subjected to multiple threats. A cost function is first formulated to convert the path planning into an optimization problem that incorporates requirements and constraints for the feasible and safe operation of the UAV. SPSO is then used to find the optimal path that minimizes the cost function by efficiently searching the configuration space of the UAV via the correspondence between the particle position and the speed, turn angle and climb/dive angle of the UAV. To evaluate the performance of SPSO, eight benchmarking scenarios have been generated from real digital elevation model maps. The results show that the proposed SPSO outperforms not only other particle swarm optimization (PSO) variants including the classic PSO, phase angle-encoded PSO and quantum-behave PSO but also other state-of-the-art metaheuristic optimization algorithms including the genetic algorithm (GA), artificial bee colony (ABC), and differential evolution (DE) in most scenarios. In addition, experiments have been conducted to demonstrate the validity of the generated paths for real UAV operations. Source code of the algorithm can be found at  \url{https://github.com/duongpm/SPSO}.
    
\end{abstract}


\end{frontmatter}

%\linenumbers

\section{Introduction}
Path planning is essential for UAVs to carry out tasks and avoid threats appearing in their operating environment. A planned path should be optimal in a specific criterion defined by the application. For most applications such as aerial photography, mapping, and surface inspection, the criterion is typically to minimize the traveling distance among the visiting locations of UAVs so that less time and fuel are required \cite{PHUNG2017,Hoang2020}. The criterion can also be maximizing the detection probability as in dynamic target search \cite{PHUNG2020}, minimizing the flight time as with surveillance and rescue \cite{Lin2009}, or finding the Pareto solution for multi-objective navigation \cite{Yin2018}. In addition, the planned path also needs to satisfy the constraints relating to safety imposed by the operating environment and feasibility imposed by the UAV. Here, safety relates to the capability of the path to guide the UAV through threats appearing in the environment such as obstacles. Feasibility involves the alignment of the path with UAV limits associated with flight time, flight altitude, fuel consumption, turning rate and climbing angle. Path planning with enhanced safety in terms of collision-free and feasible motion for UAVs therefore remains a challenging problem.

In the literature, several approaches have been proposed for UAV path planning such as graph search, cell decomposition, potential field and nature-inspired algorithms. The graph search approach splits the environment into connected discrete regions, each forms a vertex of the graph that the path is being searched. In \cite{Beard2002,McLain2005}, the Voronoi diagram has been used to generate a graph which then became the input to the Eppstein's \textit{k}-best paths algorithm \cite{Eppstein1998} to find the best path. Another graph-based algorithm is the probabilistic roadmap (PRM) that samples the configuration space of the UAV to generate vertices of the graph \cite{pettersson2006}. Similar to PRM, the rapid-exploring random trees (RRT) algorithm uses the configuration space to create a search graph. It however finds the path by recursively adding the edge that has the smallest heuristic cost to it \cite{Lin2017}. Although the graph-based algorithms are effective in generating feasible flight paths, they are not suitable to include constraints related to UAV maneuver and thus can result in large errors between the planned and flight paths.

The cell decomposition approach, on the other hand, represents the space as a grid of equal cells and employs a heuristic search to find the flight path. A* is a popular algorithm that searches the cell space using the least cost from the current location to its neighbors and the target location \cite{Hart1968,Penin2019}. This algorithm is extended in \cite{Szczerba2000} to include UAV constraints such as the turning angle. It is then modified to become bidirectional to deal with intermittent measurements \cite{Penin2019}. Cell decomposition is also used in \cite{Li2016} for path coordination between UAVs and UGVs, in \cite{Kwak2018} for flight surveillance and in \cite{Sun2015} for path prediction in real-time UAV operations. The main drawback of the cell decomposition approach however is the limitation in the scalable capacity as the number of cells exponentially increases with the search space dimension.

The potential field is another approach that directly searches the continuous space for solutions by treating the UAV as a particle moving under the influence of an artificial potential field constructed from components associated with the goal and any obstacles \cite{Barraquand1992,Jun2019}. This approach has been augmented with an additional control force to provide a shorter and smoother path \cite{Luo2015,Chen2016}. It is also combined with the Hamiltonian function to enable obstacle avoidance \cite{Hamidreza2021} or with the receding horizon optimization to obtain paths for multiple UAVs without violating the collision avoidance and network connectivity constraints \cite{Bin2015}. The potential field approach, however, does not consider the optimality of the solution. It is also known to have limitations in dealing with local minima occurred in the field.

Recently, the nature-inspired approach has become more prevalent in path planning due to its effectiveness in dealing with UAV dynamic constraints and the capacity to search for the global optimum in complex scenarios. A variety of nature-inspired algorithms have been developed for UAV path planning such as the cuckoo search \cite{PSong2020}, genetic algorithm (GE) \cite{Roberge2018,Roberge2013}, differential evolution (DE) \cite{Fu2013,ZSun2016}, artificial bee colony (ABC) \cite{XU2010535}, ant colony optimization (ACO) \cite{Yu2019}, and particle swarm optimization (PSO) \cite{PHUNG2017, Roberge2013,Fu2013,Fu2012}. Among them, PSO is commonly used with a number of variants introduced.

Inspired by the behavior of bird flocking and fish schooling, PSO is a population-based algorithm that possesses two important properties of swarm intelligence, the cognitive and social coherence \cite{Kennedy2001}. Those properties allow each particle of the swarm to search for the solution by following its own experience and the swarm experience instead of using conventional evolutionary operators like mutation and crossover. As a result, PSO is able to find the global solution with a stable convergence in a shorter computation time  compared to other nature-inspired algorithms \cite{Gaing2003}. It is also known to be less sensitive to initial conditions and variations of objective functions and is able to adapt to various environment structures via a small number of parameters including one acceleration coefficient and two weight factors \cite{Eberhart1998}. Due to its swarm nature, PSO can be parallelized to run on multiple processors, graphical processing units (GPU) or computer clusters to obtain the computation time required for both offline and online path planning \cite{lalwani2019}. Given those advantages, PSO has been widely used for path planning for mobile robot navigation with different approaches introduced such as evolutionary operator-based PSO \cite{DAS2020106312}, adaptive bare-bones PSO \cite{zhang2014adaptive} or multi-objective PSO \cite{ZHANG2013172,na2016solving}. In UAV path planning, several variants have been proposed such as the classic PSO \cite{Kennedy2001,Roberge2013}, phase angle-encoded PSO ($\theta$-PSO) \cite{wei2008,Hoang2018,Fu2012}, quantum-behaved PSO (QPSO) \cite{Sun2012,Fu2012} and discrete PSO (DPSO) \cite{PHUNG2017,clerc2004}. Those variants have the same population-based structure but differ in the way they represent the search space and the solution encoded in particles. Consequently, different solutions may have resulted under the same conditions of the operating environment, dynamic constraints, and objective function. Therefore, it is important to compare those variants in different scenarios to provide a clear insight as to which of them is preferable for UAV path planning. In addition, it is also necessary to incorporate the maneuver properties of UAVs into the algorithms to further improve their navigation capacity.  

In this study, we address the path planning problem for UAVs by first formulating an objective function that incorporates various requirements and constraints associated with the UAV and its flight path. We then introduce a new PSO algorithm that is capable of exploiting the configuration space of the UAV to generate quality solutions. For evaluation, eight scenarios have been generated with increasing levels of complexity based on the use of real digital elevation model (DEM) maps. The comparisons between SPSO and other PSO and metaheuristic algorithms are then conducted on those scenarios to evaluate their performance. In addition, experiments have been carried out to verify the feasibility of the solutions generated by SPSO for UAV operation in practical scenarios. Our contributions in this study therefore are fourfold: (i) development of a new objective function that converts the path planning into an optimization problem incorporating optimal criteria and constraints associated with the path length, threat, turn angle, climb/dive angle, and flight height for the safe and efficient operation of UAVs; (ii) proposal of a new PSO algorithm named spherical vector-based PSO (SPSO) that is capable of searching the configuration space for the global optimal solution; (iii) benchmarking the performance of PSO variants including PSO, $\theta$-PSO, QPSO and SPSO for UAV path planning; (iv) validating the generated paths for real UAV operations. 
 
The rest of this paper is structured as follows. Section~\ref{sect:formulation} introduces the steps to formulate the objective function. Section~\ref{sect:PSO} describes PSO and its variants. Section \ref{SPSO} presents SPSO and its implementation for solving the path planning problem. Section~\ref{result} provides comparison and experiment results. Finally, a conclusion is drawn to end our paper.

\section{Problem Formulation}
\label{sect:formulation}
In this study, the path planning problem is formulated via a cost function that incorporates optimal criteria and UAV constraints described as follows.

\subsection{Path optimality}
\label{optimality}
For efficient operation of UAVs, a planned path needs to be optimal in a certain criterion depending on the application. With our focus on aerial photography, mapping, and surface inspection, we choose to minimize the path length. Since the UAV is controlled via a ground control station (GCS), a flight path $X_i$ is represented as a list of $n$ waypoints that the UAV needs to fly through. Each waypoint corresponds to a path node in the search map with coordinates $P_{ij} = (x_{ij},y_{ij},z_{ij})$. By denoting the Euclidean distance between two nodes as $\left\|\overrightarrow{P_{ij}P_{i,j+1}}\right\|$, the cost $F_1$ associated to the path length can be computed as:

\begin{equation}\label{eq:cost1}
F_1(X_i) = \sum_{j=1}^{n-1} \left\|\overrightarrow{P_{ij}P_{i,j+1}}\right\|.
\end{equation}  

\subsection{Safety and feasibility constraints}
\label{safety}
\begin{figure}
	\centering
	\includegraphics[scale=1]{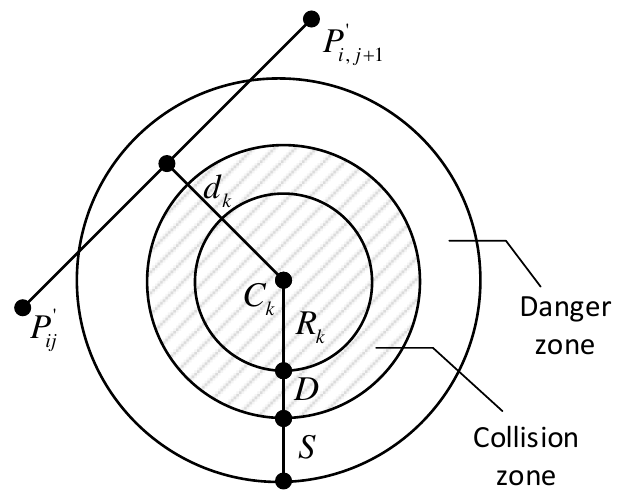}
	\caption{Determination of the threat cost.}
	\label{fig:threatcost}
\end{figure}

Apart from optimality, the planned path needs to ensure the safe operation of the UAV by guiding it through threats that are typically caused by obstacles appearing in the operation space. Let $K$ be the set of all threats, each is assumed to be prescribed in a cylinder with its projection having the center coordinate $C_k$ and radius $R_k$ as shown in Fig.\ref{fig:threatcost}. For a given path segment $\left\|\overrightarrow{P_{ij}P_{i,j+1}}\right\|$, the associated threat cost is proportional to its distance, $d_k$, to $C_k$. By considering the diameter, $D$, of the UAV and the danger distance, $S$, to the collision zone, the threat cost $F_2$ is computed across waypoints $P_{ij}$ for obstacle set $K$ as follows:

\begin{align}\label{eq:cost2}
\begin{cases}
&F_2(X_i) = \sum \limits_{j=1}^{n-1}\sum \limits_{k=1}^{K}T_k(\overrightarrow{P_{ij}P_{i,j+1}}),\\
&T_k(\overrightarrow{P_{ij}P_{i,j+1}}) = \begin{cases}
	0,  &\text{if } d_k > S + D + R_k\\
	(S + D + R_k) - d_k,    &\text{if } D + R_k < d_k \leq S + D + R_k\\
	\infty,  &\text{if } d_k \leq D + R_k.\\
	\end{cases}
\end{cases}
\end{align}

Note that while diameter $D$ is determined by the UAV size, distance $S$ depends on several factors such as the application, operating environment and positioning accuracy. For instance, $S$ can be chosen from tens of meters in static environments with good GPS signal to hundreds of meters for environments with moving objects and weak GPS signal for positioning.

\begin{figure}
	\centering
	\includegraphics[width = 0.5\textwidth]{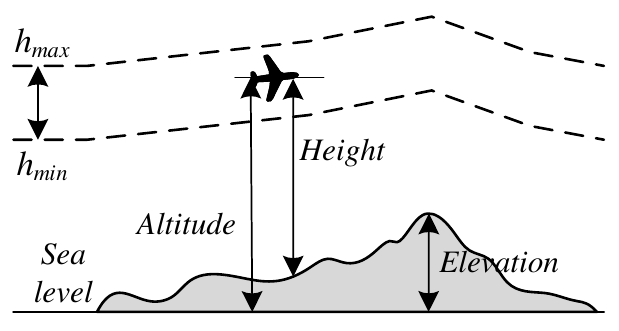}
	\caption{Altitude cost explanation.}
	\label{fig:altitudecost}
\end{figure}

During operation, the flying altitude is often constrained between the two given extrema, the minimum and maximum heights. For example with surveying and search applications, it is required the visual data to be collected by the camera at a specific resolution and field of view and thus constrain the flying altitude. Let the minimum and maximum heights to be $h_{\text{min}}$ and $h_{\text{max}}$ respectively. The altitude cost associated to a waypoint $P_{ij}$ is computed as:
\begin{align}
&H_{ij} = \begin{cases}
\vert h_{ij} - \dfrac{(h_{\text{max}} + h_{\text{min}})}{2} \vert ,  &\text{if } h_{\text{min}} \leq h_{ij} \leq h_{\text{max}}\\
\infty,  &\text{otherwise},
\end{cases}
\end{align}
where $h_{ij}$ denotes the flight height with respect to the ground as illustrated in Fig.\ref{fig:altitudecost}. It can be seen that $H_{ij}$
maintains the average height and penalises the out-of-range values. Summing $H_{ij}$ for all waypoints gives the altitude cost:

\begin{equation}\label{eq:cost3}
F_3(X_i) = \sum_{j=1}^{n} H_{ij}.
\end{equation} 

\begin{figure}
	\centering
	\includegraphics[scale=1]{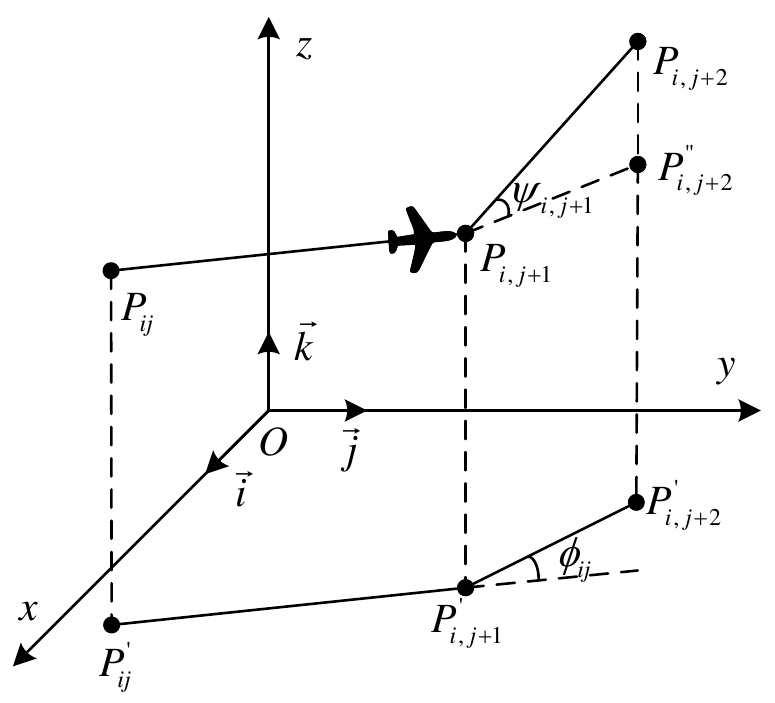}
	\caption{Turning and climbing angle calculation.}
	\label{fig:smoothcost}
\end{figure}

The smooth cost evaluates the turning and climbing rates which are essential to generate feasible paths. As shown in Fig.\ref{fig:smoothcost}, the turning angle, $\phi_{ij}$, is the angle between two consecutive path segments, $\overrightarrow{P'_{ij}P'_{i,j+1}}$ and $\overrightarrow{P'_{i,j+1}P'_{i,j+2}}$, projected on the horizontal plane $Oxy$. Let $\overrightarrow{k}$ be the unit vector in the direction of the $z$ axis, the projected vector can be calculated as:

\begin{equation}
\overrightarrow{P'_{ij}P'_{i,j+1}} = \overrightarrow{k} \times (\overrightarrow{P_{ij}P_{i,j+1}} \times \overrightarrow{k}),
\end{equation}    	
Hence, the turning angle is computed as:

\begin{equation}
\phi_{ij} = \arctan\left(\frac{\left\|\overrightarrow{P'_{ij}P'_{i,j+1}} \times \overrightarrow{P'_{i,j+1}P'_{i,j+2}}\right\|}{\overrightarrow{P'_{ij}P'_{i,j+1}} . \overrightarrow{P'_{i,j+1}P'_{i,j+2}}}\right).
\end{equation} 
The climbing angle, $\psi_{ij}$, is the angle between the path segment $\overrightarrow{P_{ij}P_{i,j+1}}$ and its projection $\overrightarrow{P'_{ij}P'_{i,j+1}}$ onto the horizontal plane. It is given by:
 
\begin{equation}
\psi_{ij} = \arctan\left(\frac{z_{i,j+1} - z_{ij}}{\left\|\overrightarrow{P'_{ij}P'_{i,j+1}}\right\|}\right).
\end{equation} 
The smooth cost is then computed as:

\begin{equation}\label{eq:cost4}
F_4(X_i) = a_1\sum_{j=1}^{n-2} \phi_{ij} + a_2\sum_{j=1}^{n-1} \mid \psi_{ij} - \psi_{i,j-1} \mid,
\end{equation}
where $a_1$ and $a_2$ are respectively the penalty coefficients of the turning and climbing angles. 

\subsection{Overall cost function}

By considering the optimality, safety and feasibility constraints associated with a path $X_i$, the overall cost function can be defined of the form: 

\begin{equation}\label{eq:cost}
	F(X_i) = \sum_{k=1}^{4} b_kF_k(X_i),
\end{equation} 
where $b_k$ is the weight coefficient, and $F_1(X_i)$ to $F_4(X_i)$ are respectively the costs associated to the path length (\ref{eq:cost1}), threat (\ref{eq:cost2}), smoothness (\ref{eq:cost3}), and flight height (\ref{eq:cost4}). The decision variable is $X_i$ including the list of $n$ waypoints $P_{ij}=(x_{ij},y_{ij},z_{ij})$ such that $P_{ij} \in O$, where $O$ is the operating space of UAVs. Given those definitions, the cost function $F$ is fully determined and can be used as the input for the path planning process.

\section{Related PSO algorithms for UAV path planning}
\label{sect:PSO}
With the cost function $F$ defined in (\ref{eq:cost}), the path planning becomes an optimization problem in which the aim is to find the path $X^*$ that minimizes $F$. As $F$ in general is a complicated multimodal function, solving it using classic methods such as hill climbing is not feasible due to local maxima. Instead, heuristic and metaheuristic methods are often used to provide quality solutions in a reasonable amount of time. This section describes the classic PSO and its variants including $\theta$-PSO and QPSO which are among the most popular metaheuristic algorithms used for UAV path planning.

\subsection{Particle swarm optimization}
PSO is a stochastic optimization method working in light of swarm intelligence. Each particle $i$ in the swarm is characterized by its position, $X_i = (x_{i1},x_{i2},...,x_{iN})$, and velocity, $V_i = (v_{i1},v_{i2},...,v_{iN})$, in the search space of $N$ dimensions. It searches for the optimal solution by compromising between its own experience reflected via the local best position, $L_i = (l_{i1},l_{i2},...,l_{iN})$, and the swarm experience reflected via the global best position, $L_g = (L_{g1},L_{g2},...,L_{gN})$. For a swarm of $M$ particles, the compromise is carried out by the following equations:

\begin{equation}
\label{eq:velocity}
v^{k+1}_{ij} \leftarrow w^kv^k_{ij} + \eta_1r_{1j}(l^k_{ij} - x^k_{ij}) + \eta_2r_{2j}(l^k_{gj} - x^k_{ij})
\end{equation} 

\begin{equation}
\label{eq:position}
x^{k+1}_{ij} \leftarrow x^k_{ij} + v^{k+1}_{ij}, (i=1,2,...,M; j=1,2,...,N),
\end{equation}
where $k$ represents the $k$th generation, $x_{ij} \in [x_{min},x_{max}]$ and $v_{ij} \in [v_{min},v_{max}]$ are respectively the $j$th dimension of the $i$th particle's position and velocity, $w^k$ is the inertial weight, $\eta_1$ and $\eta_2$ are respectively the cognitive and social coefficients, and $r_{1j}$ and $r_{2j}$ are two random samples within $[0,1]$ drawn from a uniform probability distribution. The values of $\eta_1$ and $\eta_2$ determine the moving tendency of particles toward the local best and global best position. 
The weight $w^k$, on the other hand, represents the compromise between the exploration (global search) and exploitation (local search). It is often chosen to be smaller over generations to increase the exploitation when the swarm is getting closer to the optimal solution.

When using PSO for UAV path planning, the position of each particle encodes a candidate path. Hence, the swarm is equivalent to a matrix of $M$ paths, $X = [X_1,X_2,...,X_M]^T$, each includes a list of $N$ waypoints of the form:

\begin{equation}
X_i = (x_{i1},y_{i1},z_{i1},x_{i2}, y_{i2}, z_{i2},...,x_{i,N}, y_{i,N}, z_{i,N}).
\end{equation}

As the start and end points of all paths are fixed, they are not included in the particle position. A path of $n$ waypoints is thus represented by a particle of dimension $3N$, $N= n-2$. During the optimization process, the particles evolve according to (\ref{eq:velocity}) and (\ref{eq:position}) based on the evaluation in (\ref{eq:cost}) to converge to the best path.

\subsection{Phase angle-encoded particle swarm optimization}
$\theta$-PSO uses angles instead of Cartesian coordinates to represent particle positions \cite{wei2008,Hoang2018,Fu2012}. In $\theta$-PSO, a path of $n$ waypoints is described by a vector of $3N$ angles:

\begin{equation}
\Theta_i = (\theta_{i1},...,\theta_{iN},\theta_{i,N+1},...,\theta_{i,2N},\theta_{i,2N+1},...,\theta_{i,3N}),
\end{equation}
where $N = n-2$ and $\theta_{ij}$ is within the interval $[-\pi/2,\pi/2]$. The velocity associated to each particle is then expressed by angle increments as:

\begin{equation}
\Delta\Theta_i = (\Delta\theta_{i1},...,\Delta\theta_{iN},\Delta\theta_{i,N+1},...,\Delta\theta_{i,2N},\Delta\theta_{i,2N+1},...,\Delta\theta_{i,3N}).
\end{equation}
Hence, the update equations for $\theta$-PSO are given by:

\begin{equation}
\label{eq:velocity}
\Delta\theta^{k+1}_{ij} \leftarrow w^k\Delta\theta^k_{ij} + \eta_1r_{1j}(\gamma^k_{ij} - \theta^k_{ij}) + \eta_2r_{2j}(\gamma^k_{gj} - \theta^k_{ij})
\end{equation} 

\begin{equation}
\label{eq:position}
\theta^{k+1}_{ij} \leftarrow \theta^k_{ij} + \Delta\theta^{k+1}_{ij}, (i=1,2,...,M; j=1,2,...,3N),
\end{equation}  
where $\Gamma_i = [\gamma_{i1},\gamma_{i2},...,\gamma_{i,3N}]$ and $\Gamma_g = [\gamma_{g1},\gamma_{g2},...,\gamma_{g,3N}]$ are respectively the phase angles of the local and global best positions of particle $i$. 

To evaluate the fitness, a monotonic function is used to map particles from the angular space to the coordinate space. Let the monotonic function be $f:[-\pi/2,\pi/2] \rightarrow [x_{min},x_{max}]$, there is one and only one position $X_i$ mapped by $f$ corresponding to any given position $\Theta_i$. That position is given by \cite{wei2008}:
   
\begin{align}
\label{eq:decode}
\begin{cases}
&x_{ij} = f(\theta_{ij}),\\
&f(\theta_{ij})= \dfrac{1}{2} {[(x_{max} - x_{min})sin(\theta_{ij}) + x_{max} + x_{min}]}.\\
\end{cases}
\end{align}

It can be seen from (\ref{eq:decode}) that $\theta$-PSO introduces nonlinearity to the candidate paths by adding more waypoints to their middle section and thus aims to improve the search capacity in that area of the operating environment.

\subsection{Quantum-behaved particle swarm optimization}
QPSO assumes particles have a quantum state described by a wavefunction $\psi(x,t)$ and is attracted by a Delta potential well \cite{Sun2012}. The probability that a particle appears at position $x$ is then described via its probability density function $\vert\psi(x,t)\vert^2$, which can be derived from the time-dependent Schr\"{o}dinger equation. By using the Monte Carlo simulation method, that position is updated by the following equation:

\begin{equation}
\label{eq:QPSO}
x^{k+1}_{ij} = p^k_{ij} \pm 0.5L^k_{ij}ln(1/r),
\end{equation}
where $r \in (0,1)$ is a random number drawn from a uniform probability distribution and $p^k_{ij}$ is a local attractor computed as:

\begin{equation}
p^{k}_{ij} = al^k_{ij} + (1-a)l^k_{gj},
\end{equation} 
where $a \in (0,1)$ is a random number of uniform distribution. $L^k_{ij}$ is the parameter computed by:
\begin{align}
L^k_{ij} = 2\beta\vert mbest^k_j - x^k_{ij} \vert, \\
mbest^k_j = \sum_{i=1}^{M}l^k_{ij}/M,
\end{align} 
where $mbest$ is the mean best position of the swarm and $\beta$ is the contraction-expansion coefficient. Noting that particles in QPSO do not maintain the velocity component but only the position.

In path planning, QPSO represents a flight path as a set of $N$ waypoints similar to PSO. Those waypoints are encoded through the position of particles which is then updated by \ref{eq:QPSO}.

\section{Spherical vector-based PSO for UAV path planning}
Exploiting maneuver characteristics of UAVs, we propose in this study the SPSO algorithm and provide its implementation to solve the path planning problem. 
 
\label{SPSO}
\subsection{Spherical vector-based PSO algorithm}
SPSO encodes each path as a set of vectors, each describes the movement of the UAV from one waypoint to another. Those vectors are represented in the spherical coordinate system with three components including the magnitude $\rho \in (0,path\_length)$, elevation angle $\psi \in (-\pi/2,\pi/2)$, and azimuth angle $\phi \in (-\pi,\pi)$. A flight path $\Omega_i$ with $N$ nodes is then represented by a hyper spherical vector of $3N$ dimensions:

\begin{equation}
\Omega_i = (\rho_{i1},\psi_{i1}, \phi_{i1}, \rho_{i2},\psi_{i2},\phi_{i2},..., \rho_{iN},\psi_{iN}, \phi_{iN}), N = n-2.
\end{equation}  
By describing the position of a particle as $\Omega_i$, the velocity associated to that particle is described by an incremental vector:

\begin{equation}
\Delta \Omega_i = (\Delta\rho_{i1},\Delta\psi_{i1}, \Delta\phi_{i1}, \Delta\rho_{i2},\Delta\psi_{i2},\Delta\phi_{i2},..., \Delta\rho_{iN},\Delta\psi_{iN}, \Delta\phi_{iN}).
\end{equation} 
Denoting spherical vector $(\rho_{ij},\psi_{ij},\phi_{ij})$ as $u_{ij}$ and velocity $(\Delta\rho_{ij},\Delta\psi_{ij},\Delta\phi_{ij})$ as $\Delta u_{ij}$, the update equations for SPSO are given by:

\begin{equation}
\label{eq:SPSOvelocity}
\Delta u^{k+1}_{ij} \leftarrow w^k\Delta u^k_{ij} + \eta_1r_{1j}(q^k_{ij} - u^k_{ij}) + \eta_2r_{2j}(q^k_{gj} - u^k_{ij})
\end{equation} 

\begin{equation}
\label{eq:SPSOposition}
u^{k+1}_{ij} \leftarrow u^k_{ij} + \Delta u^{k+1}_{ij}, (i=1,2,...,M; j=1,2,...,N),
\end{equation}
where $Q_i = (q_{i1},q_{i2},...,q_{i,N})$ and $Q_g = (q_{g1},q_{g2},...,q_{g,N})$ are respectively the sets of vectors representing the local and global best positions of particle $i$. 

In order to determine $Q_i$ and $Q_g$, it is required to map a vector-based flight path $\Omega_i$ to a direct path $X_i$ so that the associated cost can be evaluated. The mapping of vector $u_{ij}=(\rho_{ij},\psi_{ij},\phi_{ij}) \in \Omega_i$ to waypoint $P_{ij} = (x_{ij},y_{ij},z_{ij}) \in X_i$ can be conducted as:

\begin{align} \label{eq:decode1}
&x_{ij} = x_{i,j-1} + \rho_{ij}sin\psi_{ij}cos\phi_{ij},\\
\label{eq:decode2}
&y_{ij} = y_{i,j-1} + \rho_{ij}sin\psi_{ij}sin\phi_{ij},\\
\label{eq:decode3}
&z_{ij} = z_{i,j-1} + \rho_{ij}cos\psi_{ij}.
\end{align}
Denoting the map as $\xi:\Omega \rightarrow X$, the local and global best positions can be computed as:

\begin{equation} \label{eq:LBest}
Q_i = \left\{\begin{array}{ll}
\Omega_i \quad \text{if $F(\xi(\Omega_i)) < F(\xi(Q_{i-1}))$}\\
Q_{i-1} \quad \mathrm{otherwise}
\end{array}\right., 
\end{equation}

\begin{equation} \label{eq:GBest}
Q_g = \underset{Q_i}{\operatorname{argmin}}F(\xi(Q_i)). 
\end{equation}

The rationale for the use of spherical vectors in SPSO is to achieve safety enhancement of navigation via the interrelationship between the magnitude, elevation and azimuth components of the vectors with the speed, turning angle and climbing angle of the UAV. As a result, the particles of SPSO search for solutions in the configuration space instead of the Cartesian space to reach a higher probability of finding quality solutions. More importantly, constraints relating to the turning and climbing angles can be directly implemented via the elevation and azimuth angles of the spherical vector so that the search space can be significantly reduced. In some scenarios, for example when the UAV flies at a constant speed, the magnitude can be fixed to further reduce the search space for extending the search capacity.

\begin{algorithm}
	\caption{Pseudo code of SPSO for UAV path planning.}
	\label{fig:pseudocode}
	\tcc{Initialization:}
	Get search map and initial path planning information \;
	Set swarm parameters $w$, $\eta_1$, $\eta_2$, $swarm\_size$\; 
	\ForEach {particle $i$ in swarm} {
		\quad	Create a random path $\Omega^0_i$\;
		\quad	Assign $\Omega^0_i$ to particle's position\;
		\quad	Compute $fitness$ $F(\xi(\Omega_i))$ of the particle\;
		\quad	Set $local\_best$ $Q_i$ of the particle to its $fitness$\; 
	}
	Set $global\_best$ $Q_g$ to the best fit particle\;
	\tcc{Evolutions:}
	\For{$k \gets 1$ to $max\_generation$} {
		\ForEach {particle $i$ in swarm} { 
			Compute velocity $\Delta\Omega^k_i$; \tcc*[f]{Eq.\ref{eq:SPSOvelocity}} \\
			Compute new position $\Omega^k_i$; 
			\tcc*[f]{Eq.\ref{eq:SPSOposition}} \\
			Map $\Omega^k_i$ to $X^k_i$ in Cartesian space;  \tcc*[f]{Eq.\ref{eq:decode1} - \ref{eq:decode3}} \\
			Update $fitness$ $F(X^k_i)$;  
			\tcc*[f]{Eq.\ref{eq:cost}} \\
			Update $local\_best$ $Q_i$;  
			\tcc*[f]{Eq.\ref{eq:LBest}} \\
		}
		Update $global\_best$ $Q_g$; 
		\tcc*[f]{Eq.\ref{eq:GBest}} \\
		Save $best\_position$ $\Omega^*$ associated with $Q_g$ ;  \tcc*[f]{the best path} \\
	}
\end{algorithm}

\subsection{Implementation of SPSO for UAV path planning}
The pseudo code of SPSO is shown in Fig.\ref{fig:pseudocode}. It shares the same structure as other PSO variants including parameter initialization, particle generation and swarm evolution, but differs from others in the representation of particles' position, velocity and update equations. Therefore, parallelism can be used as adopted in \cite{PHUNG2017} to speed up the calculation process. During algorithm execution, infeasible solutions appeared will be assigned an infinite cost value so that they will excluded from the final output solutions.
	
\section{Results} \label{result}
To evaluate the performance of SPSO, we have conducted a number of comparisons and experiments with details as follows.

\subsection{Scenario setup}
\begin{figure*}
	
	\begin{subfigure}{0.5\textwidth}
		\centering
		\includegraphics[width=\textwidth]{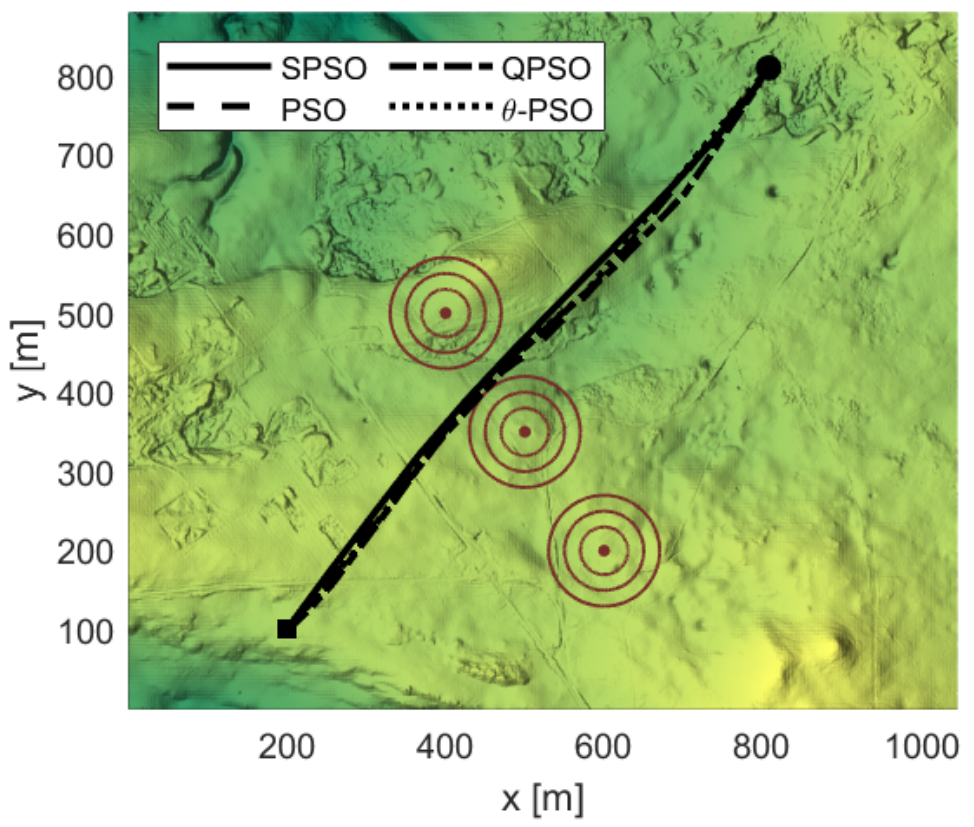}
		\caption{Scenario 1}
		\label{fig:toppath1}
	\end{subfigure}%
	\begin{subfigure}{0.5\textwidth}
		\centering
		\includegraphics[width=\textwidth]{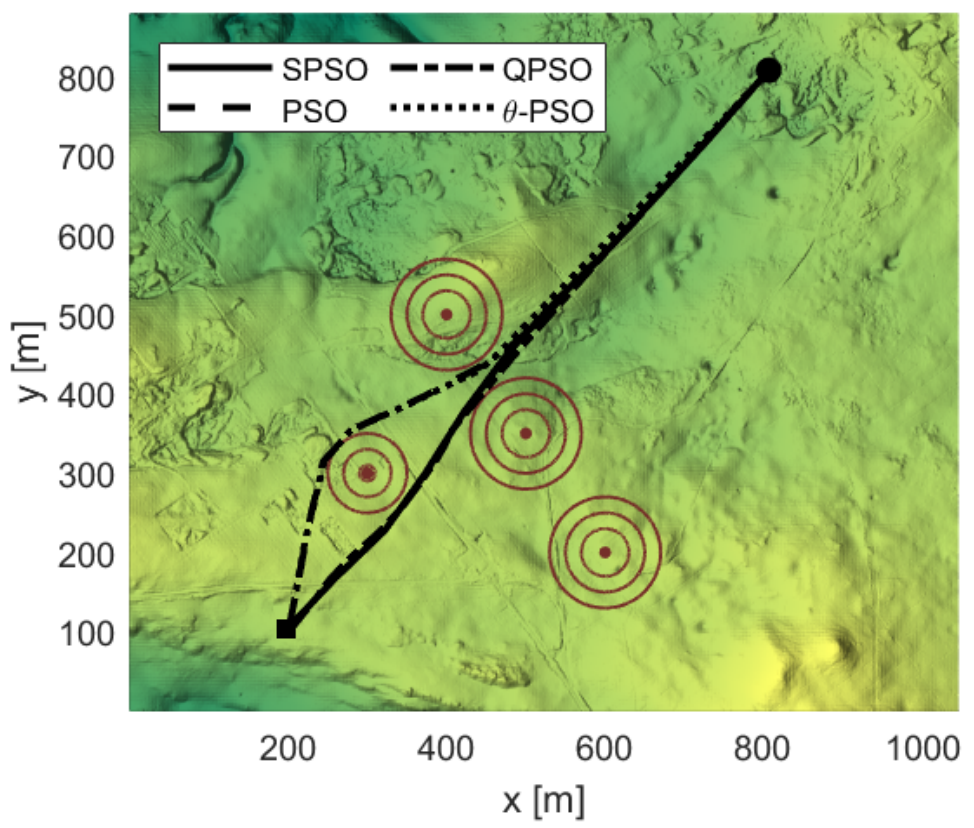}
		\caption{Scenario 2}
		\label{fig:toppath2}
	\end{subfigure}
	\begin{subfigure}{0.5\textwidth}
		\centering
		\includegraphics[width=\textwidth]{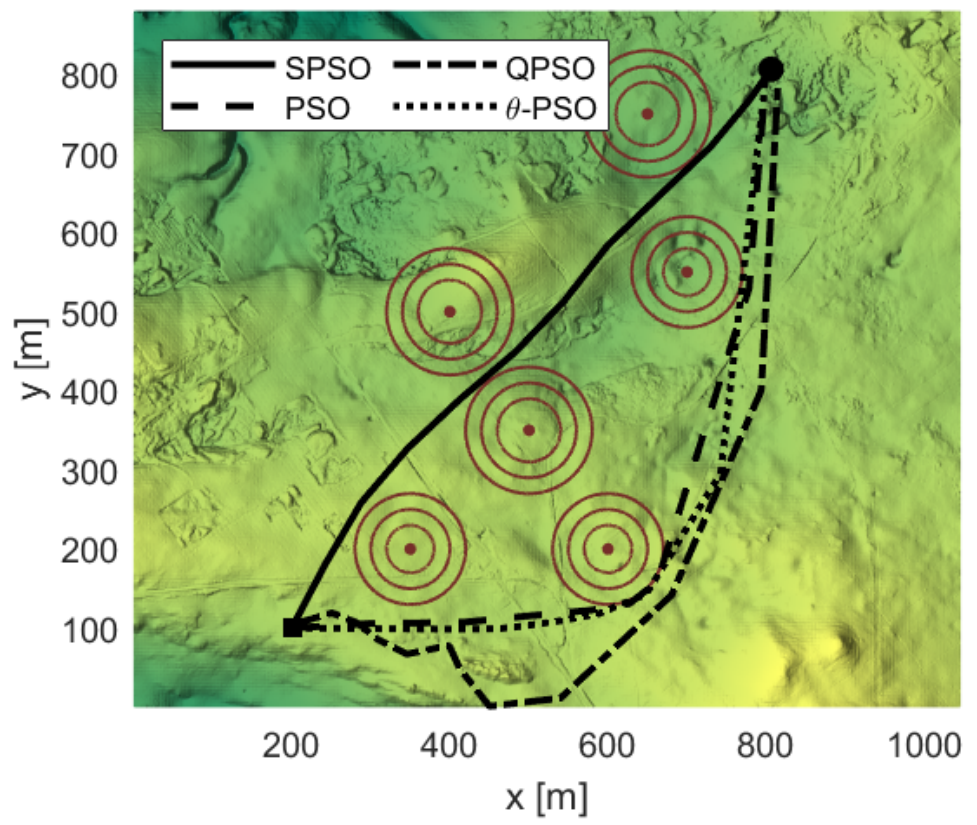}
		\caption{Scenario 3}
		\label{fig:toppath3}
	\end{subfigure}%
	\begin{subfigure}{0.5\textwidth}
		\centering
		\includegraphics[width=\textwidth]{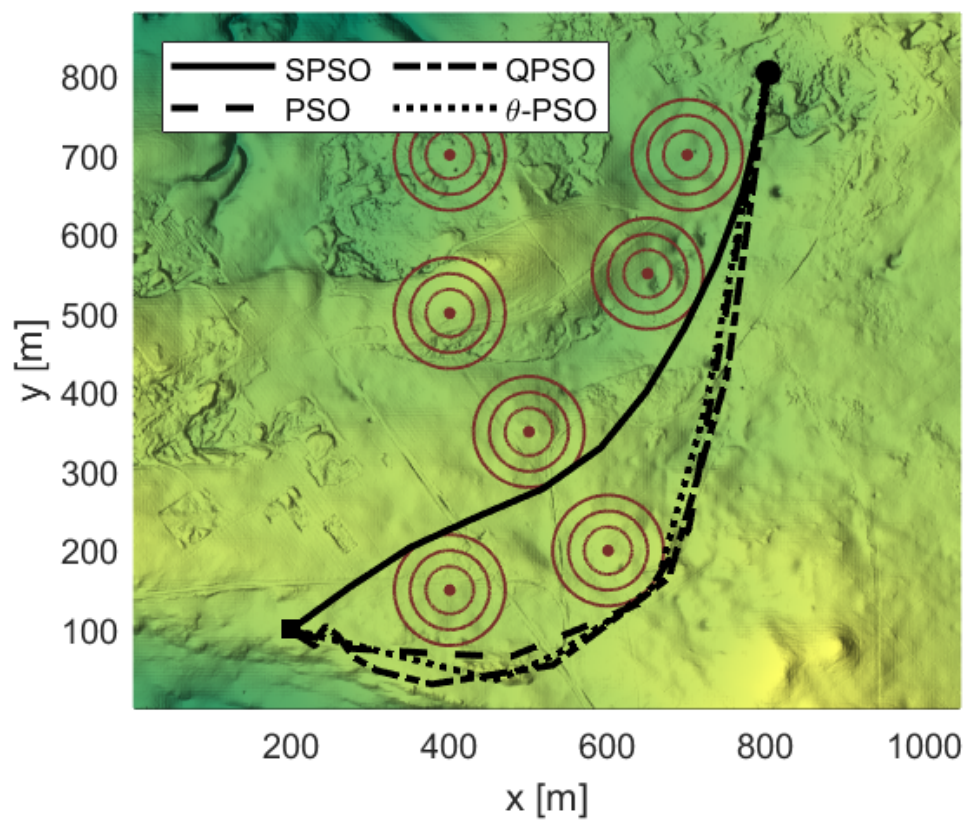}
		\caption{Scenario 4}
		\label{fig:toppath4}
	\end{subfigure}
	
	\centering
	\caption{Top view of the paths generated by the PSO variants for scenarios 1 to 4}
	\label{fig:toppath14}
\end{figure*}

\begin{figure*}
	
	\begin{subfigure}{0.5\textwidth}
		\centering
		\includegraphics[width=\textwidth]{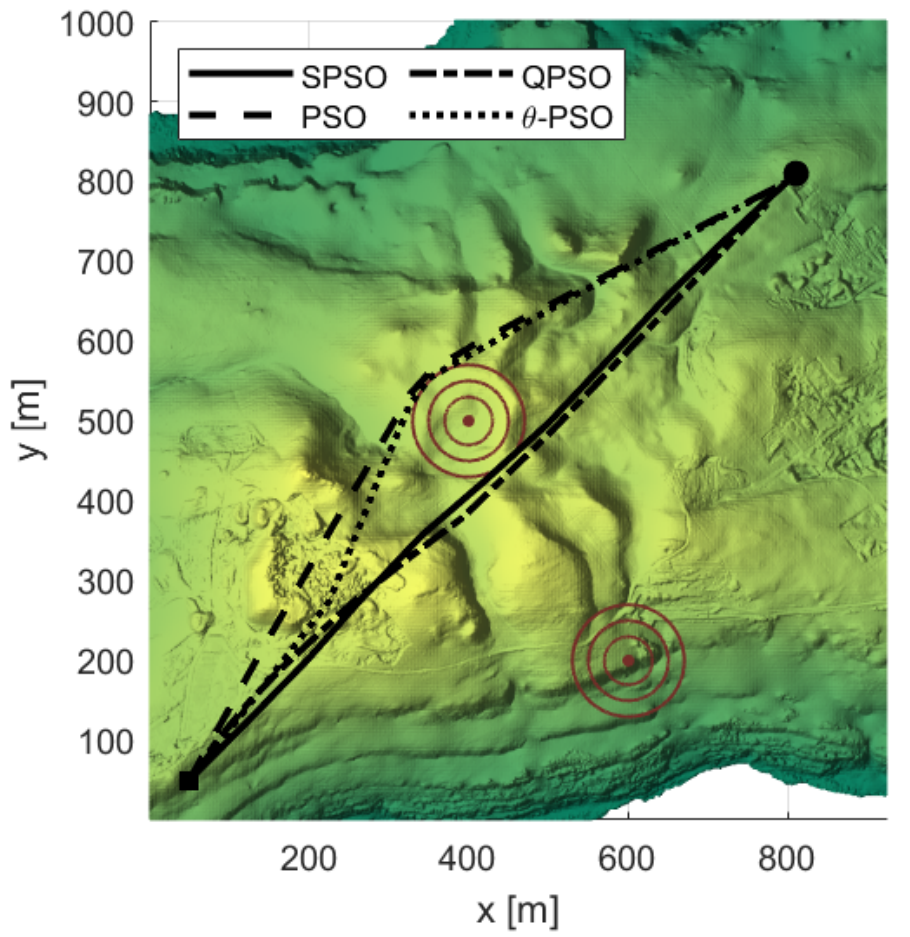}
		\caption{Scenario 5}
		\label{fig:toppath5}
	\end{subfigure}%
	\begin{subfigure}{0.5\textwidth}
		\centering
		\includegraphics[width=\textwidth]{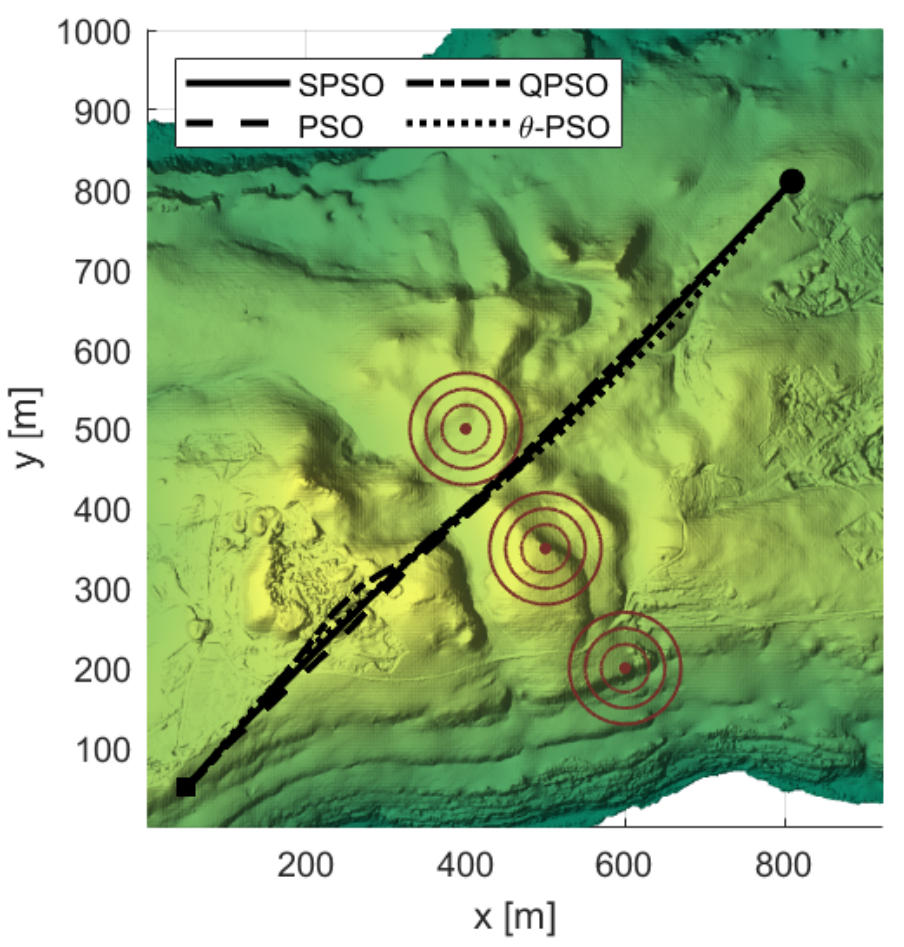}
		\caption{Scenario 6}
		\label{fig:toppath6}
	\end{subfigure}
	\begin{subfigure}{0.5\textwidth}
		\centering
		\includegraphics[width=\textwidth]{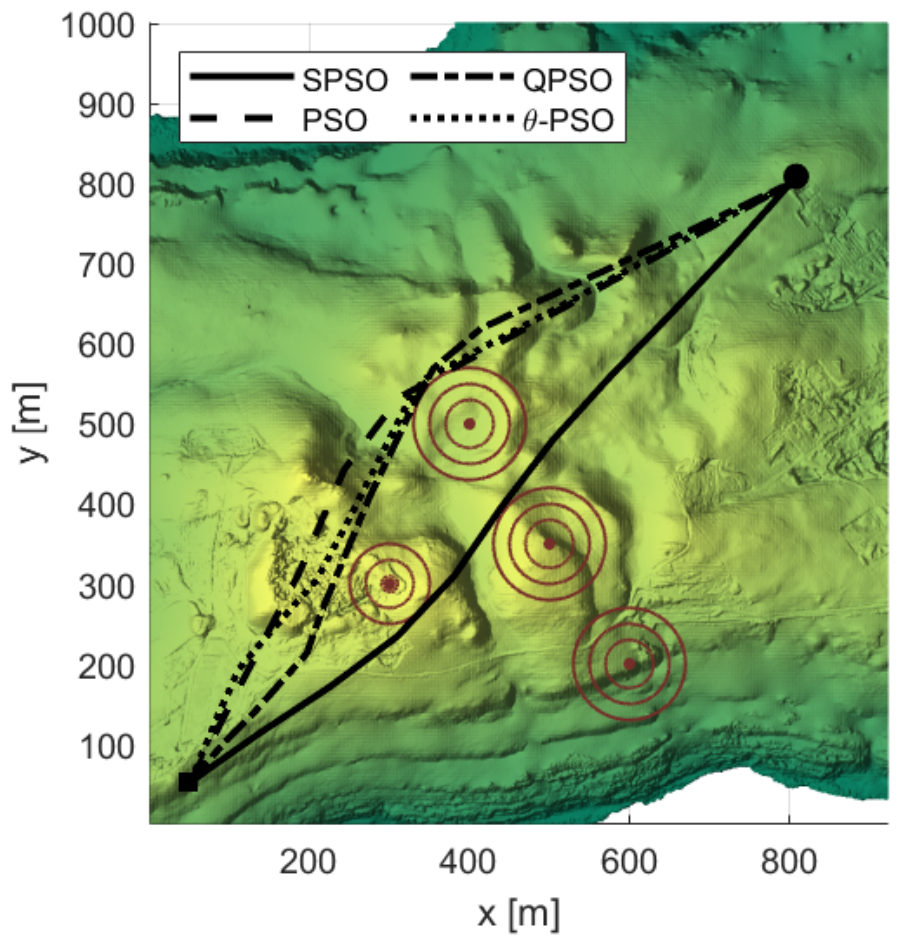}
		\caption{Scenario 7}
		\label{fig:toppath7}
	\end{subfigure}%
	\begin{subfigure}{0.5\textwidth}
		\centering
		\includegraphics[width=\textwidth]{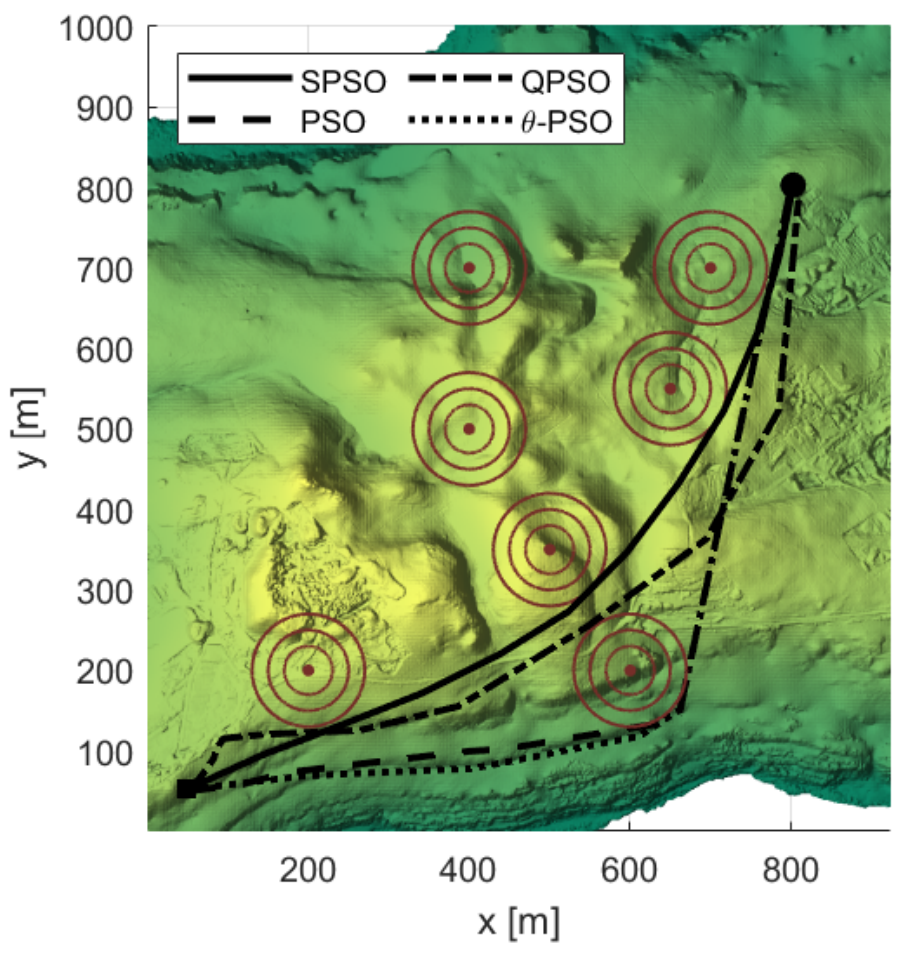}
		\caption{Scenario 8}
		\label{fig:toppath8}
	\end{subfigure}
	
	\centering
	\caption{Top view of the paths generated by the PSO variants for scenarios 5 to 8}
	\label{fig:toppath58}
\end{figure*}

The scenarios used for evaluation are based on real digital elevation model (DEM) maps derived from LiDAR sensors \cite{australia2015digital}. Two areas of Christmas Island in Australia with different terrain structures are selected and then augmented to generate eight benchmarking scenarios as shown in Fig.\ref{fig:toppath14} and Fig. \ref{fig:toppath58}. In those scenarios, the number and location of threats, represented as red cylinders, are chosen at different levels of complexity.

For comparisons, all PSO variants are implemented with the same set of parameters: $w = 1$ with the damping rate of 0.98, $\eta_1 = 1.5$ and $\eta_2 = 1.5$. The swarm size is chosen to be 500 particles and the number of iterations is 200. The number of waypoints are respectively selected as $n=12$ and $n=22$ corresponding to 10 and 20 line segments. In each comparison, all algorithms are run 10 times to find the average and standard deviation values. In addition, a statistical metric named paired sample $t$-test \cite{Hsu2005} is used to evaluate the significance of mean differences between SPSO and other PSO algorithms. The notation $D+$ implies that the mean value of SPSO is statistically better than the compared PSO, $D-$ implies the opposite, whereas $N$ means that the difference is insignificant and $NA$ stands for ``Not Applicable''. The level of confidence in $t$-test evaluations is set to $\alpha = 0.05$ equivalent to $95\%$.

\subsection{Comparison between PSO algorithms}
The top view of the resultant paths for $n=12$ generated by PSO algorithms are shown in Fig.\ref{fig:toppath14} and Fig.\ref{fig:toppath58}. It can be seen that all algorithms are able to generate feasible paths that fulfill the requirements on the path length, threat, turn angle, climb/dive angle and height. Their optimality, however, varies with scenarios. For simple scenarios 1, 2, 5, and 6, all algorithms converge well with slight differences in their fitness values. The $t$-test values in Table 1 also show that some differences in scenarios 1 and 5 are statistically insignificant. For more complicated scenarios 3, 4, 7, and 8, their performance however is much different. SPSO is able to obtain near-optimal solutions, whereas PSO and $\theta$-PSO only converge to relatively good solutions. QPSO is not able to find quality solutions. This result can be further confirmed by Table \ref{tab:fitness1} which presents the average, standard deviation, and paired sample $t$-test of the fitness values. It shows that SPSO statistically achieves the best fitness with $D+$ $t$-test in most scenarios while QPSO is only good in simple scenarios. PSO and $\theta$-PSO introduce relatively good results in all scenarios with stable convergence reflected via small deviations.   

\begin{table*}
	\centering
	\caption{Fitness values of the paths generated by the PSO variants with 10 line segments ($n=12$)}
	\label{tab:fitness1}
	\begin{tabular}{cllllllllllll}
		\hline
		\rule{0pt}{3ex}
		Scenario & \multicolumn{3}{c}{SPSO} & \multicolumn{3}{c}{PSO} & \multicolumn{3}{c}{$\theta$-PSO} & \multicolumn{3}{c}{QPSO} \\
		-        & Mean&Std&$t$-test   & Mean&Std&$t$-test    & Mean&Std&$t$-test            & Mean&Std&$t$-test            \\
		\hline
		1        & 4683&104&NA   & 4683&98&$N$    &\textbf{4643}&50&$N$            & 4826&162&$D+$            \\
		2        & \textbf{4699}&94&NA    & 5059&41&$D+$           & 5006&69&$D+$            & 5958&220&$D+$            \\
		3        & \textbf{5486}&38&NA    & 5761&20&$D+$           & 5766&32&$D+$             & 7470&462&$D+$            \\
		4        & \textbf{4994}&28&NA   & 5781&56&$D+$           & 5794&46&$D+$            & 7120&761&$D+$            \\
		5        & \textbf{5441}&27&NA            & 5476&37&$N$           & 5518&37&$D+$            & 5508&33&$D+$    \\
		6        & \textbf{5362}&59&NA   & 5514&67&$D+$            & 5486&45&$D+$            & 5474&21&$D+$           \\
		7        & \textbf{5778}&94&NA   & 5838&39&$D+$            & 5800&43&$N$            & 5965&193&$D+$           \\
		8        & \textbf{6006}&63&NA   & 6396&29&$D+$            & 6368&46&$D+$            & 8093&259&$D+$           \\
		\bottomrule
	\end{tabular}
\end{table*}
 
\begin{figure*}
	
	\begin{subfigure}{0.5\textwidth}
		\centering
		\includegraphics[width=\textwidth]{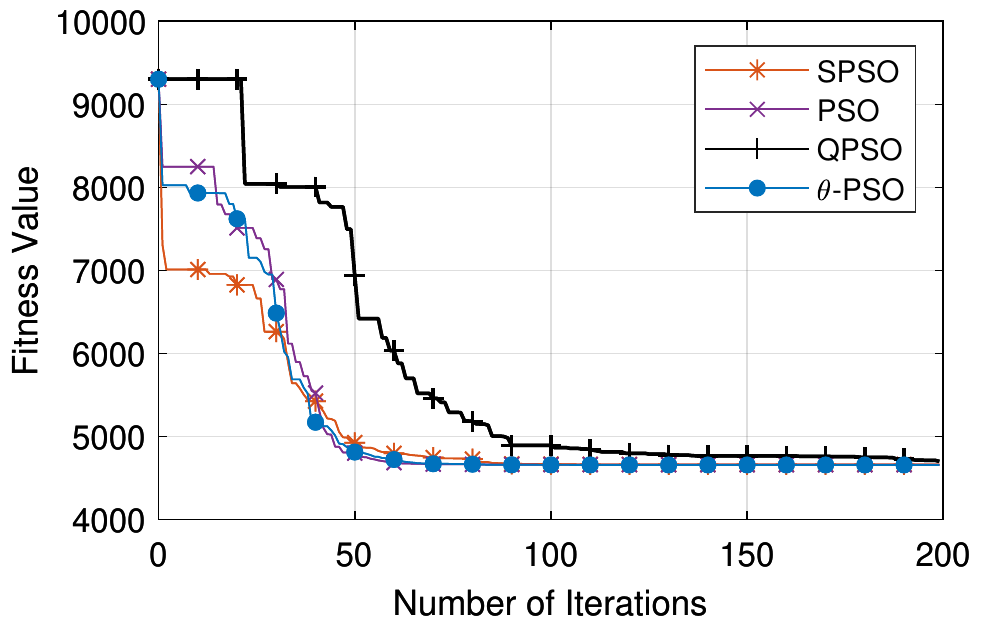}
		\caption{Scenario 1}
		\label{fig:best1}
	\end{subfigure}%
	\begin{subfigure}{0.5\textwidth}
		\centering
		\includegraphics[width=\textwidth]{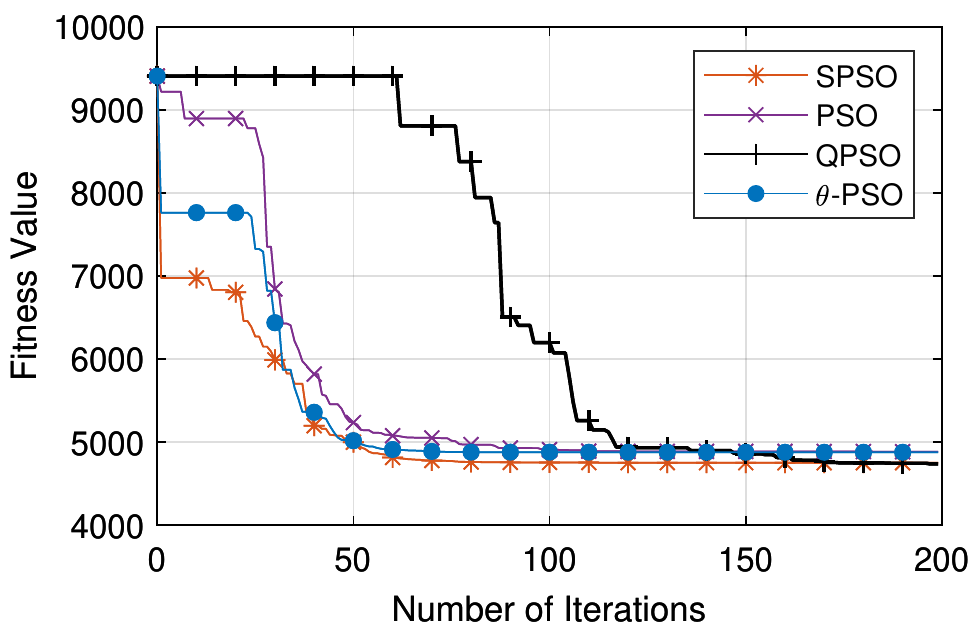}
		\caption{Scenario 2}
		\label{fig:best2}
	\end{subfigure}
	\begin{subfigure}{0.5\textwidth}
		\centering
		\includegraphics[width=\textwidth]{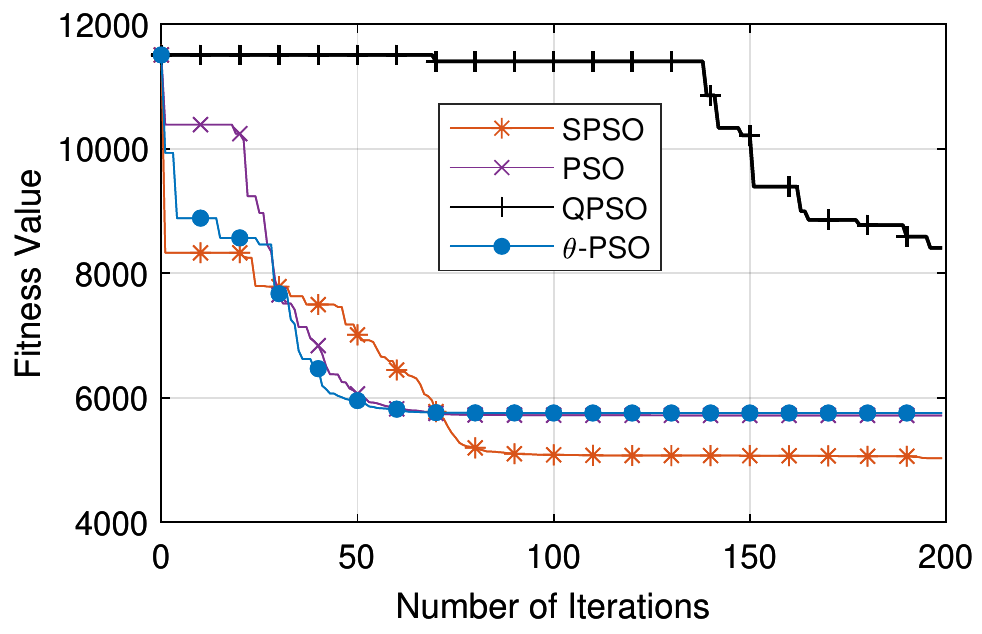}
		\caption{Scenario 3}
		\label{fig:best3}
	\end{subfigure}%
	\begin{subfigure}{0.5\textwidth}
		\centering
		\includegraphics[width=\textwidth]{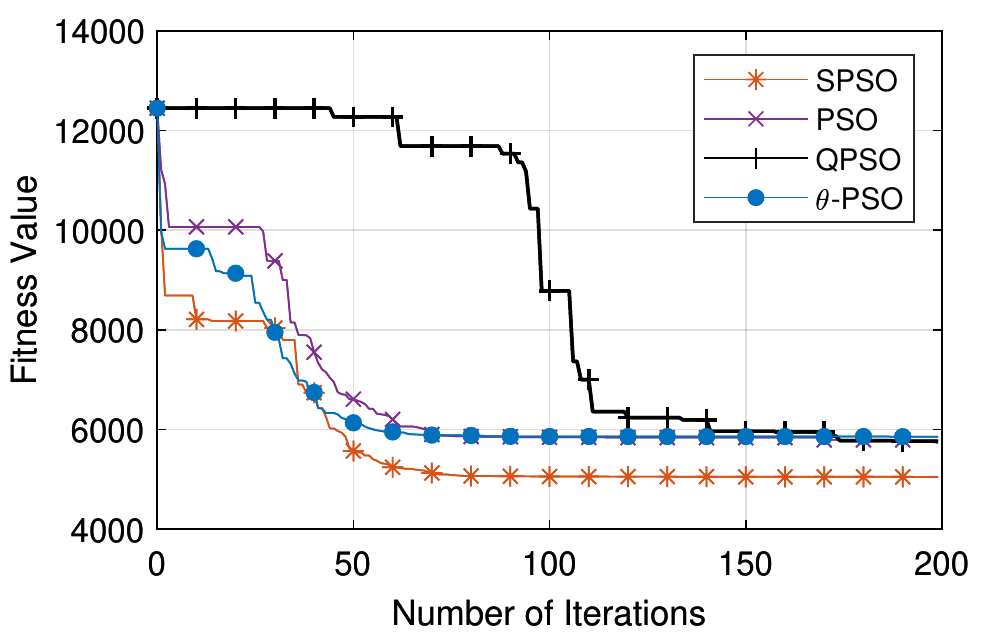}
		\caption{Scenario 4}
		\label{fig:best4}
	\end{subfigure}
	\begin{subfigure}{0.5\textwidth}
		\centering
		\includegraphics[width=\textwidth]{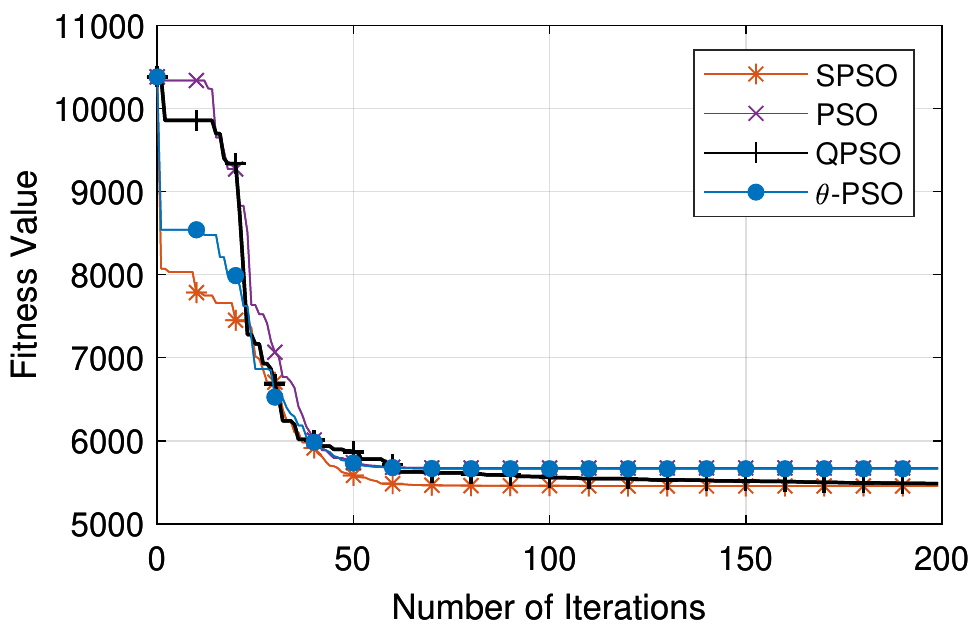}
		\caption{Scenario 5}
		\label{fig:best5}
	\end{subfigure}
	\begin{subfigure}{0.5\textwidth}
		\centering
		\includegraphics[width=\textwidth]{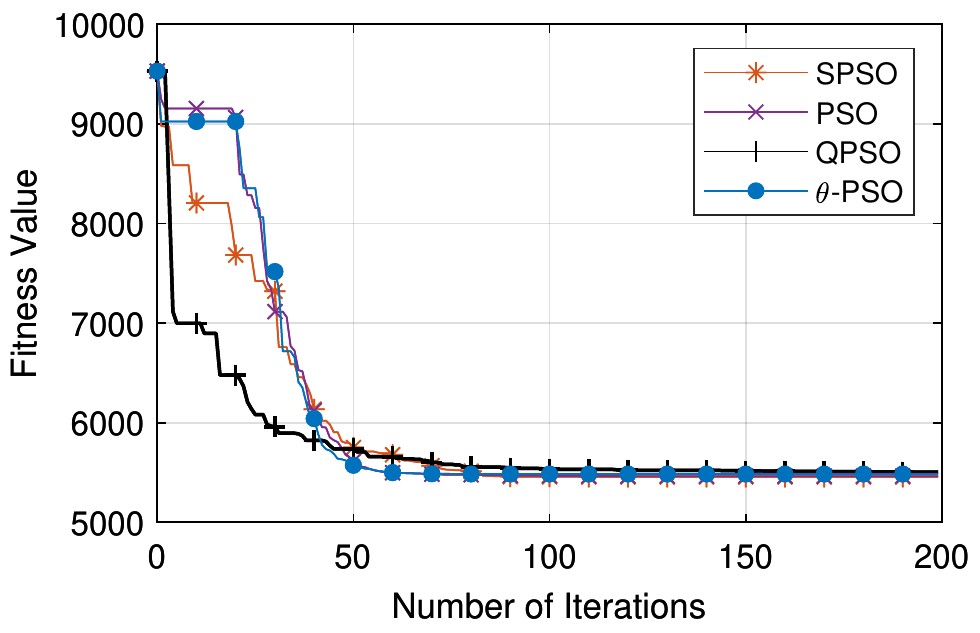}
		\caption{Scenario 6}
		\label{fig:best6}
	\end{subfigure}		
	\begin{subfigure}{0.5\textwidth}
		\centering
		\includegraphics[width=\textwidth]{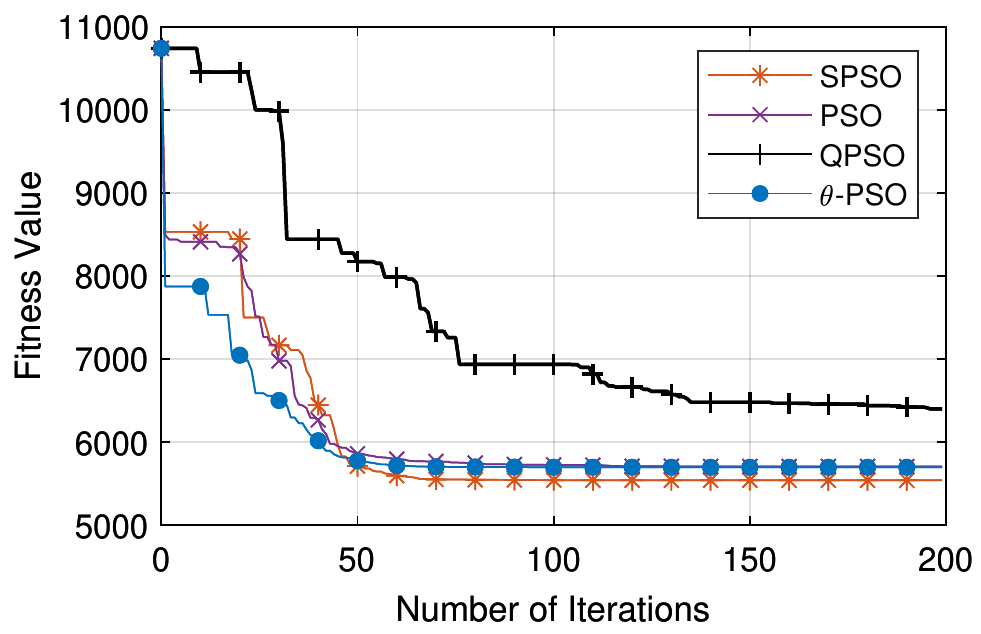}
		\caption{Scenario 7}
		\label{fig:best7}
	\end{subfigure}
	\begin{subfigure}{0.5\textwidth}
		\centering
		\includegraphics[width=\textwidth]{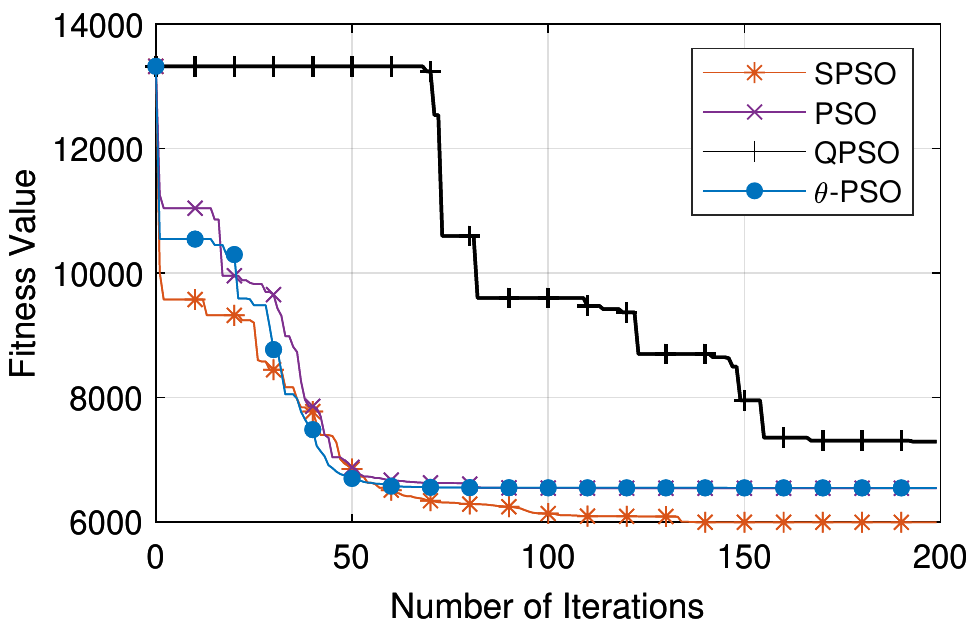}
		\caption{Scenario 8}
		\label{fig:best8}
	\end{subfigure}	
	
	\centering
	\caption{Best fitness values over iterations of the PSO algorithms}
	\label{fig:bestCost}
\end{figure*}

\begin{figure*}
	
	\begin{subfigure}{0.5\textwidth}
		\centering
		\includegraphics[width=\textwidth]{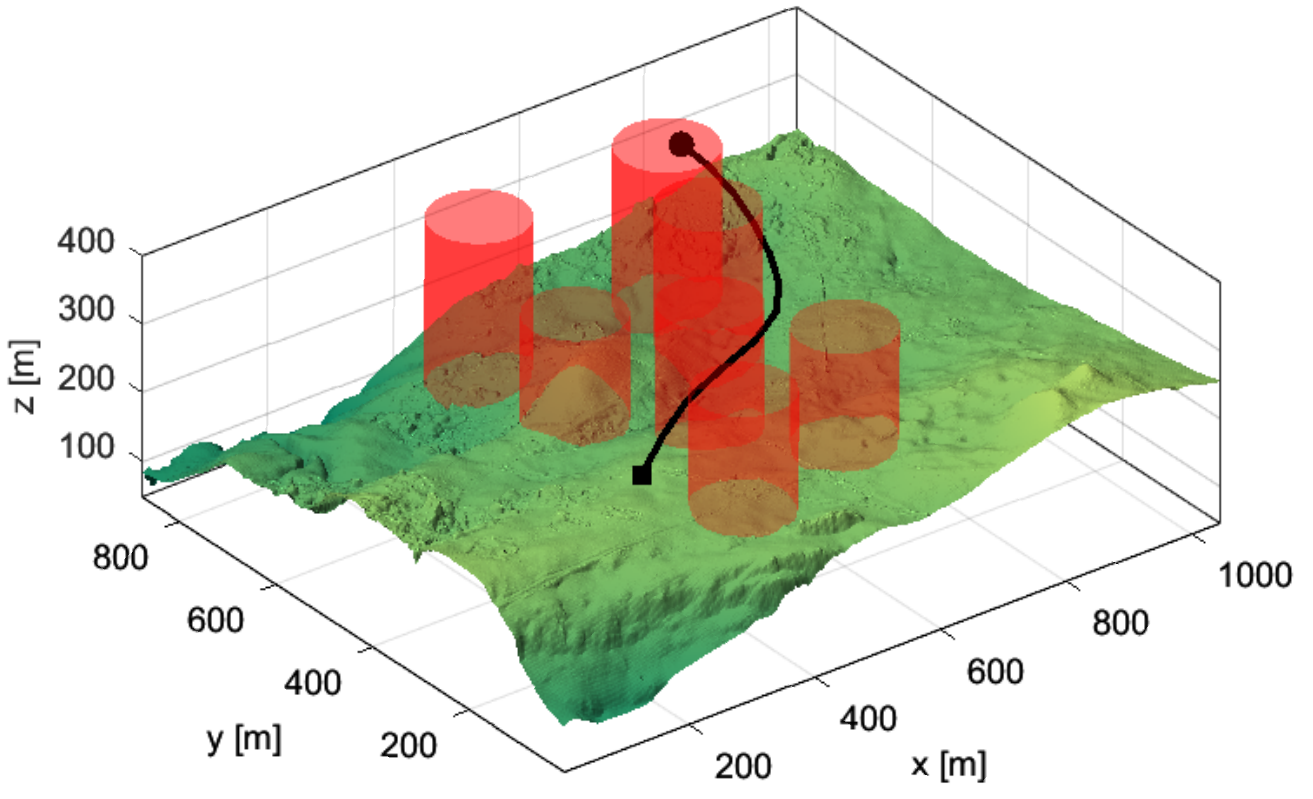}
		\caption{Scenario 4: 3D view}
		\label{fig:path1}
	\end{subfigure}%
	\begin{subfigure}{0.5\textwidth}
		\centering
		\includegraphics[width=\textwidth]{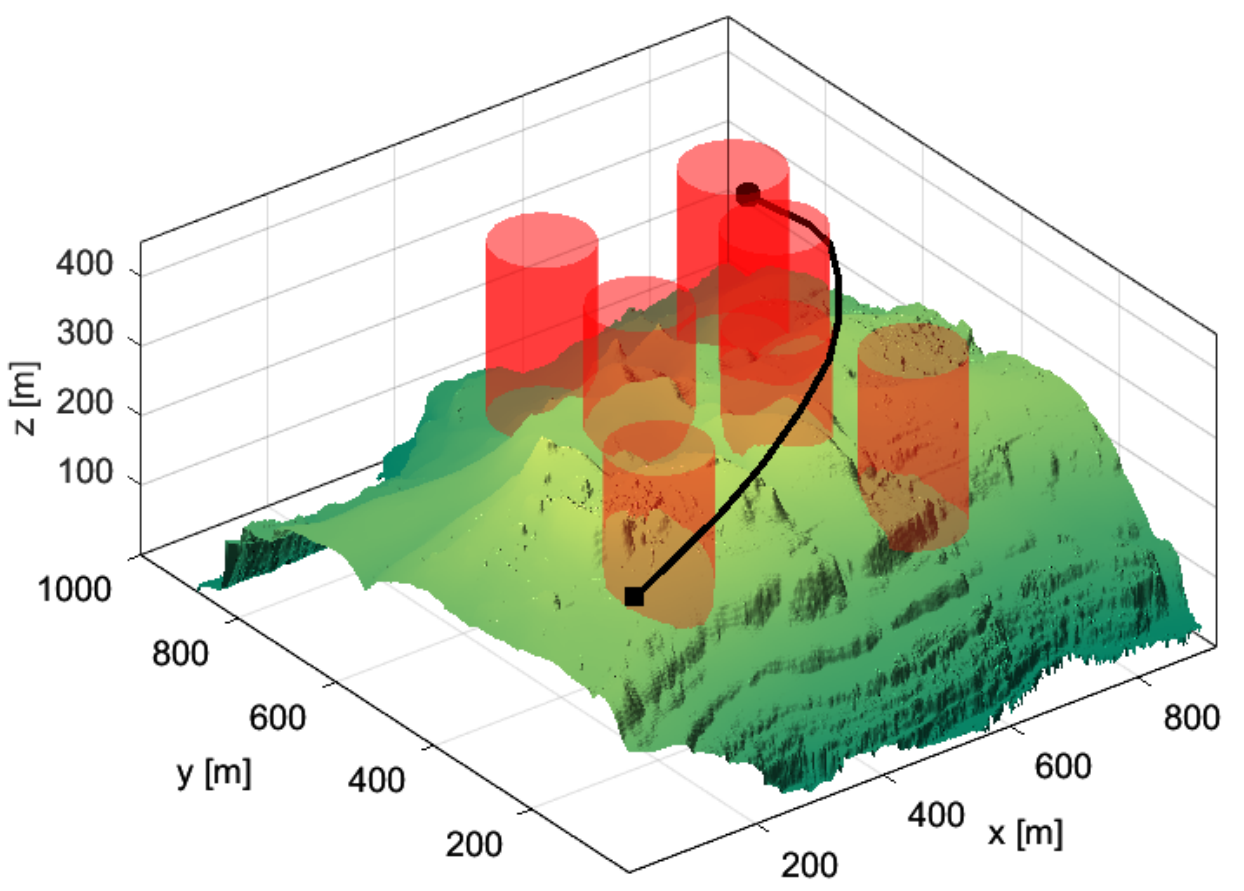}
		\caption{Scenario 8: 3D view}
		\label{fig:path2}
	\end{subfigure}
	\begin{subfigure}{0.5\textwidth}
		\centering
		\includegraphics[width=\textwidth]{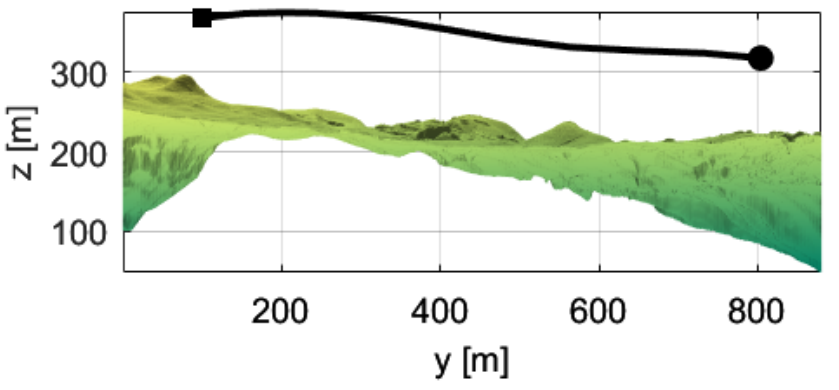}
		\caption{Scenario 4: Side view}
		\label{fig:path3}
	\end{subfigure}%
	\begin{subfigure}{0.5\textwidth}
		\centering
		\includegraphics[width=\textwidth]{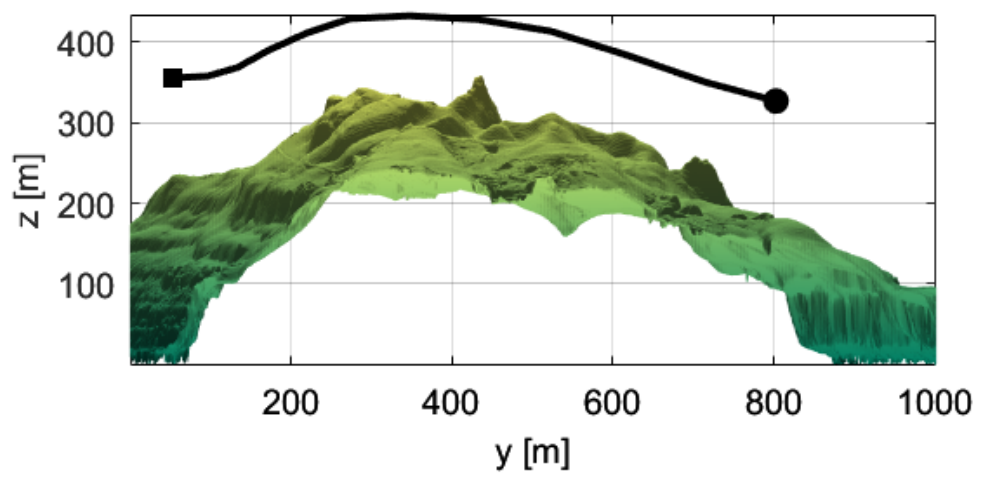}
		\caption{Scenario 8: Side view}
		\label{fig:path4}
	\end{subfigure}
	
	\centering
	\caption{The planned paths generated by SPSO for scenarios 4 and 8}
	\label{fig:path}
\end{figure*}

Figure \ref{fig:bestCost} provides a closer look at the behavior of the variants by showing their best fitness over iterations. It is recognizable that all variants converge in a similar fashion except QPSO. It is due to the fact that QPSO does not originate from the interaction of biological swarms but the transition in quantum states of particles. On another note, SPSO presents the best performance as it has a direct mapping between the properties of particles and UAV parameters to gain advantages in exploring the search space. Figure \ref{fig:path} shows the 3D and side views of the paths obtained by SPSO for scenarios 4 and 8, the two most challenging scenarios. It can be seen that the paths are smooth and valid with the flight height maintained properly with respect to the terrain. 

To compare the scalability of PSO variants, we increased the number of waypoints to $n = 22$ in another comparison. The result presented in Table \ref{tab:fitness2} shows that QPSO does not perform well in most scenarios, especially scenarios 3, 4 and 8 when its particles could not evolve to find better solutions. PSO and $\theta$-PSO perform properly for simple scenarios but for complicated scenarios like 4, 5, and 8, the quality of solutions is degraded due to their limitation in exploring large search space. SPSO, on the other hand, performs well in most scenarios thanks to the spherical vector-based encoding mechanism that allows its particles to search in the configuration space. 

\begin{table*}
	\centering
	\caption{Fitness values of the paths generated by the PSO variants with 20 line segments ($n=22$)}
	\label{tab:fitness2}
	\begin{tabular}{cllllllllllll}
		\hline
		\rule{0pt}{3ex}
		Scenario & \multicolumn{3}{c}{SPSO} & \multicolumn{3}{c}{PSO} & \multicolumn{3}{c}{$\theta$-PSO} & \multicolumn{3}{c}{QPSO} \\
		-        & Mean&Std&$t$-test   & Mean&Std&$t$-test    & Mean&Std&$t$-test            & Mean&Std&$t$-test            \\
		\hline
		1        & \textbf{4757}&91&NA   & 4821&54&$N$    & 4827&63&$N$            & 5246&236&$D+$            \\
		2        & 4906&141&NA    & \textbf{4900}&162&$N$           & 4903&136&$N$            & 5379&201&$D+$            \\
		3        & \textbf{6202}&172&NA    & 6521&291&$D+$           & 6260&355&$N$             & 16926&0&$D+$            \\
		4        & \textbf{5373}&179&NA   & 6146&335&$D+$           & 6201&451&$D+$            & 15406&354&$D+$            \\
		5        & \textbf{5806}&222&NA            & 6188&133&$D+$           & 6330&237&$D+$            & 6572&320&$D+$    \\
		6        & \textbf{5718}&130&NA   & 5844&275&$D+$            & 5733&173&$N$            & 6049&68&$D+$           \\
		7        & \textbf{5951}&132&NA   & 5992&197&$N$            & 6338&363&$D+$            & 7143&372&$D+$           \\
		8        & \textbf{6152}&111&NA   & 7016&281&$D+$            & 7134&782&$D+$            & 13828&0&$D+$           \\
		\bottomrule
	\end{tabular}
\end{table*}

\begin{table*}
	\centering
	\caption{Fitness values of the paths generated by the SPSO and other metaheuristic algorithms with 10 line segments ($n=12$)}
	\label{tab:fitness3}
	\begin{tabular}{cllllllllllll}
		\hline
		\rule{0pt}{3ex}
		Scenario & \multicolumn{3}{c}{SPSO} & \multicolumn{3}{c}{GA} & \multicolumn{3}{c}{DE} & \multicolumn{3}{c}{ABC} \\
		-        & Mean&Std&$t$-test   & Mean&Std&$t$-test    & Mean&Std&$t$-test            & Mean&Std&$t$-test            \\
		\hline
		1        & \textbf{4683}&104&NA   & 4782
		&145&$N$    & 5014&6&$D+$            & 4822&49&$D+$            \\
		2        & \textbf{4699}&94&NA    & 5357&113&$D+$           & 5040&14&$D+$            & 5020&56&$D+$            \\
		3        & \textbf{5486}&38&NA    & 6761&94&$D+$           & 5716&2&$D+$             & 5882&266&$D+$            \\
		4        & \textbf{4994}&28&NA   & 6325&224&$D+$           & 5741&4&$D+$           & 5325&118&$D+$            \\
		5        & \textbf{5441}&27&NA            & 5676&117&$D+$           & 5482&9&$D+$            & 5608&34&$D+$    \\
		6        & \textbf{5362}&59&NA   & 5424&81&$D+$            & 5665&36&$D+$            & 5676&41&$D+$           \\
		7        & 5778&94&NA   & 5919&75&$D+$            & \textbf{5633}&17&$D-$            & 5976&75&$D+$           \\
		8        & \textbf{6006}&63&NA   & 7274&554&$D+$            & 6290&72&$D+$            & 6719&90&$D+$           \\
		\bottomrule
	\end{tabular}
\end{table*}

\subsection{Comparison with other metaheuristic algorithms}
To further evaluate the performance of SPSO, we have compared its performance with other state-of-the-art metaheuristic algorithms including the genetic algorithm (GA), artificial bee colony (ABC), and differential evolution (DE). GA is implemented as in \cite{Roberge2013} with three mutation operations: add a node, delete a node and merge two nodes. ABC is implemented in its standard form \cite{KARABOGA2008687}. DE is also implemented in its standard form \cite{storn1997}, but with some changes in the swarm size and iterations due to its characteristic which performs better over a large number of iterations \cite{LI2020106193}. Specifically, DE is implemented with the swarm size of 100 and the iteration number of 1000  to ensure the algorithm convergence at the same number of fitness evaluations as with other algorithms. 

Table \ref{tab:fitness3} shows the fitness results. It can be seen that SPSO outperforms other algorithms with increasing margins and $D+$ $t$-test for complicated scenarios 3, 4, 7 and 8. However, DE performs well for complicated scenarios and has a stable convergence with small deviations due to its exploration capacity over a large number of iterations. ABC performs relatively well while GA shows the least stable performance. The reason for the unstable performance of GA lies in its operator `delete a node'. This operator removes waypoints from a path to reduce the search dimensions so that the probability of finding quality solutions is increased. That operator, however, also reduces the resolution of candidate paths causing them insufficient to adapt to complex threats as with scenarios 3, 4, and 8. 

\begin{figure*}
	
	\begin{subfigure}{0.5\textwidth}
		\centering
		\includegraphics[width=\textwidth]{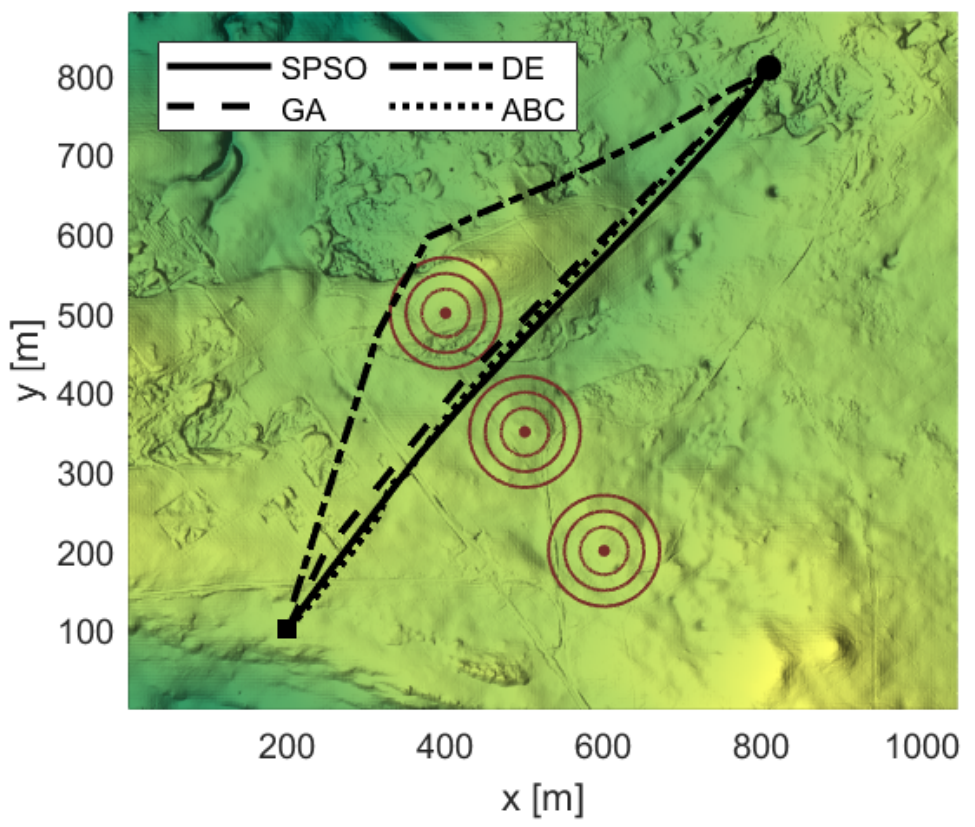}
		\caption{Scenario 1}
		\label{fig:toppath1b}
	\end{subfigure}%
	\begin{subfigure}{0.5\textwidth}
		\centering
		\includegraphics[width=\textwidth]{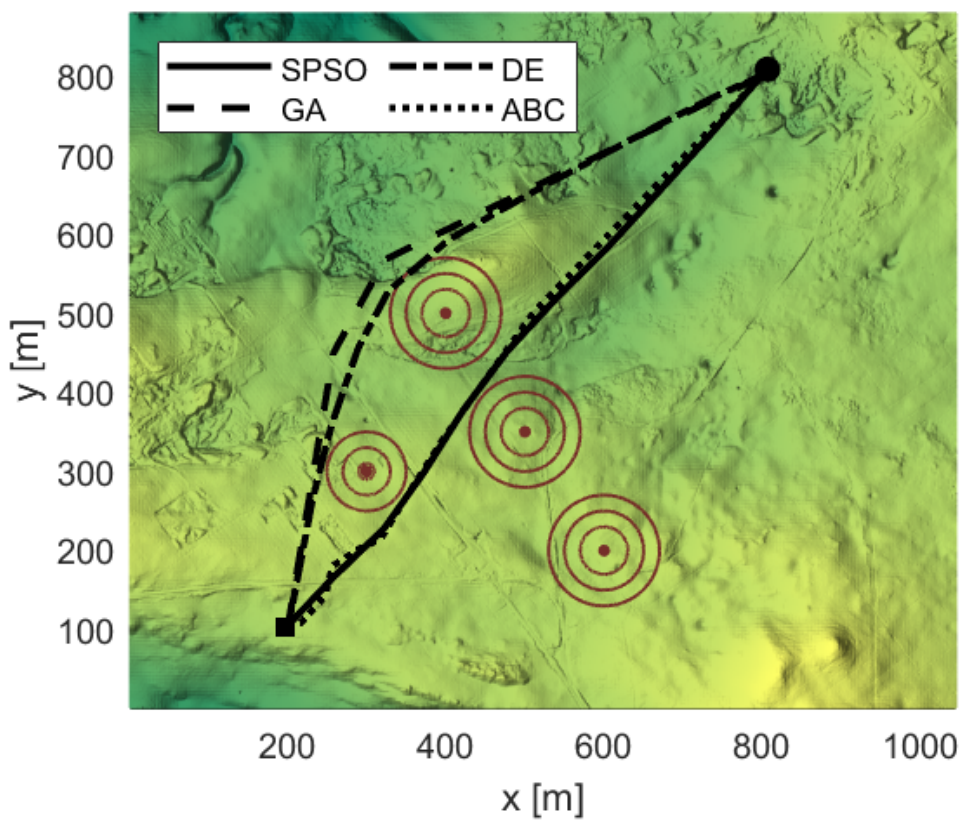}
		\caption{Scenario 2}
		\label{fig:toppath2b}
	\end{subfigure}
	\begin{subfigure}{0.5\textwidth}
		\centering
		\includegraphics[width=\textwidth]{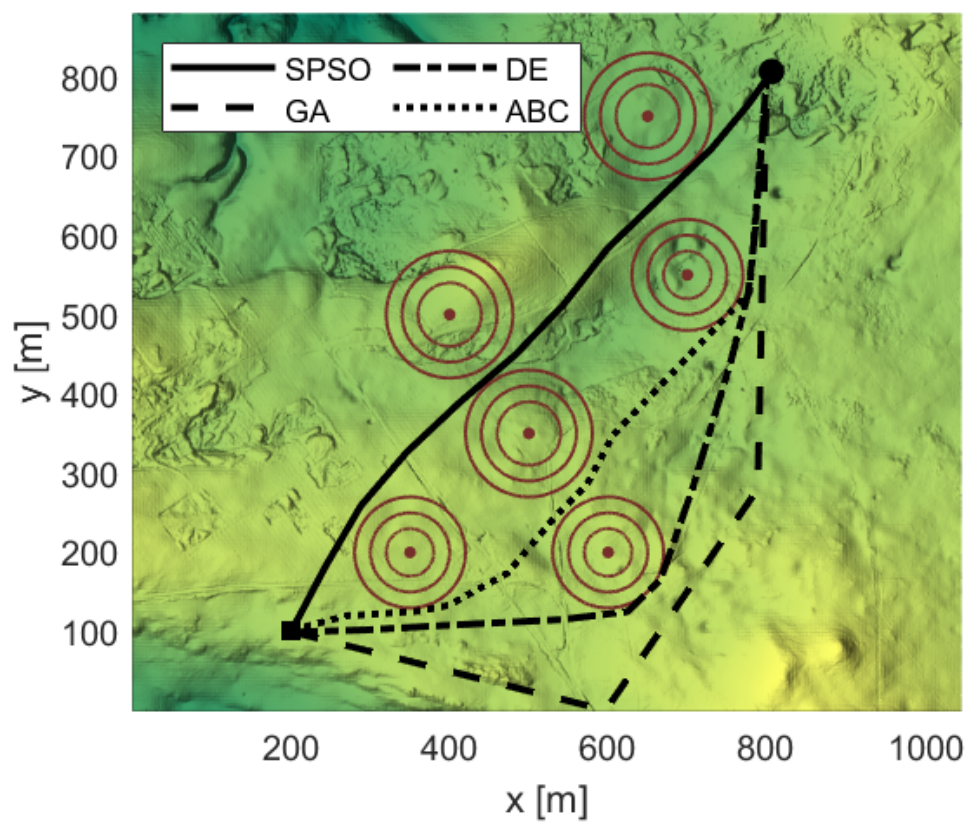}
		\caption{Scenario 3}
		\label{fig:toppath3b}
	\end{subfigure}%
	\begin{subfigure}{0.5\textwidth}
		\centering
		\includegraphics[width=\textwidth]{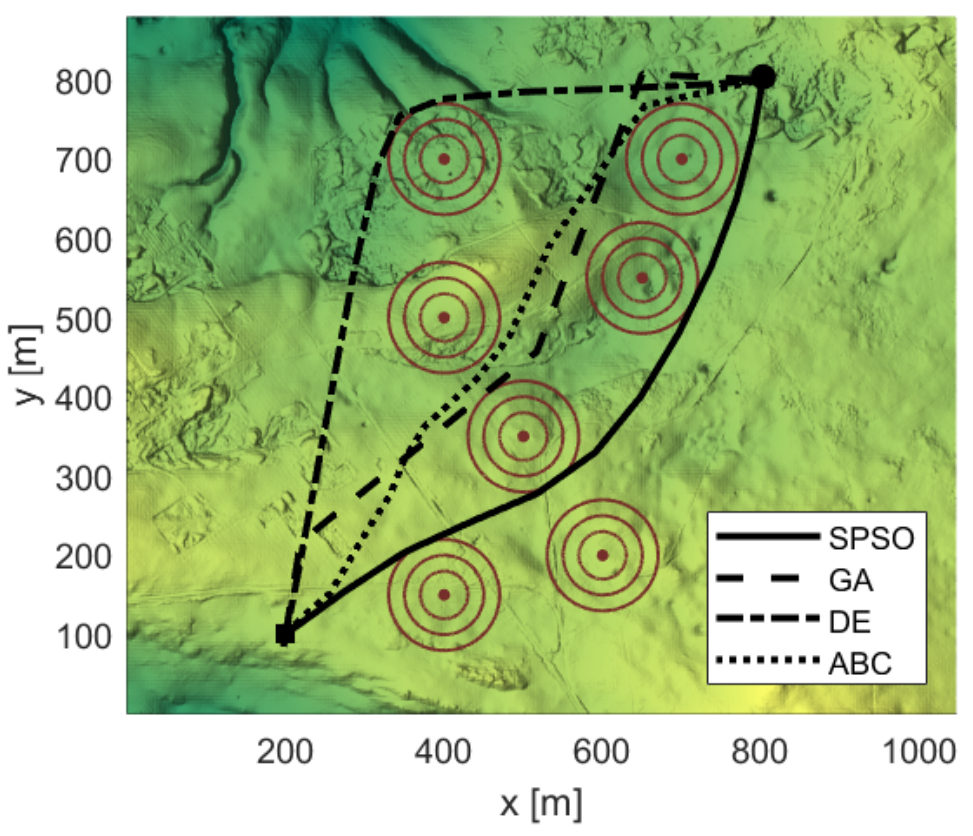}
		\caption{Scenario 4}
		\label{fig:toppath4b}
	\end{subfigure}
	
	\centering
	\caption{Top view of the paths generated by SPSO and other metaheuristic algorithms for scenarios 1 to 4}
	\label{fig:toppath14b}
\end{figure*}

\begin{figure*}
	
	\begin{subfigure}{0.5\textwidth}
		\centering
		\includegraphics[width=\textwidth]{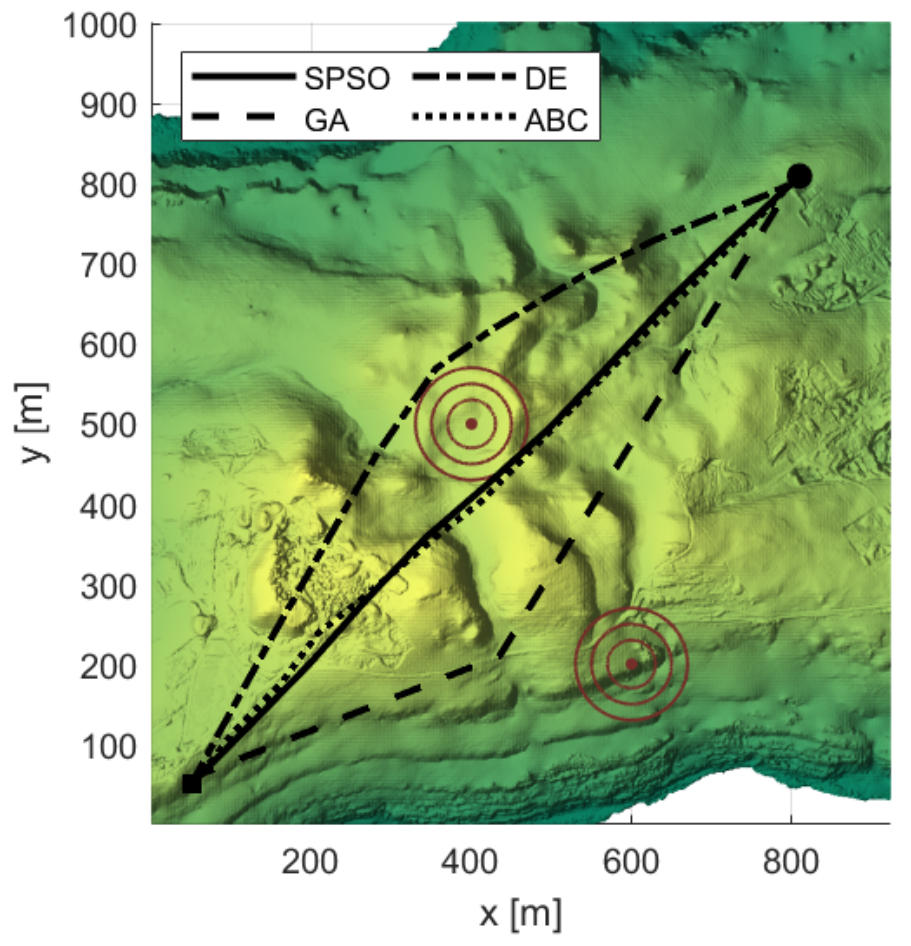}
		\caption{Scenario 5}
		\label{fig:toppath5b}
	\end{subfigure}%
	\begin{subfigure}{0.5\textwidth}
		\centering
		\includegraphics[width=\textwidth]{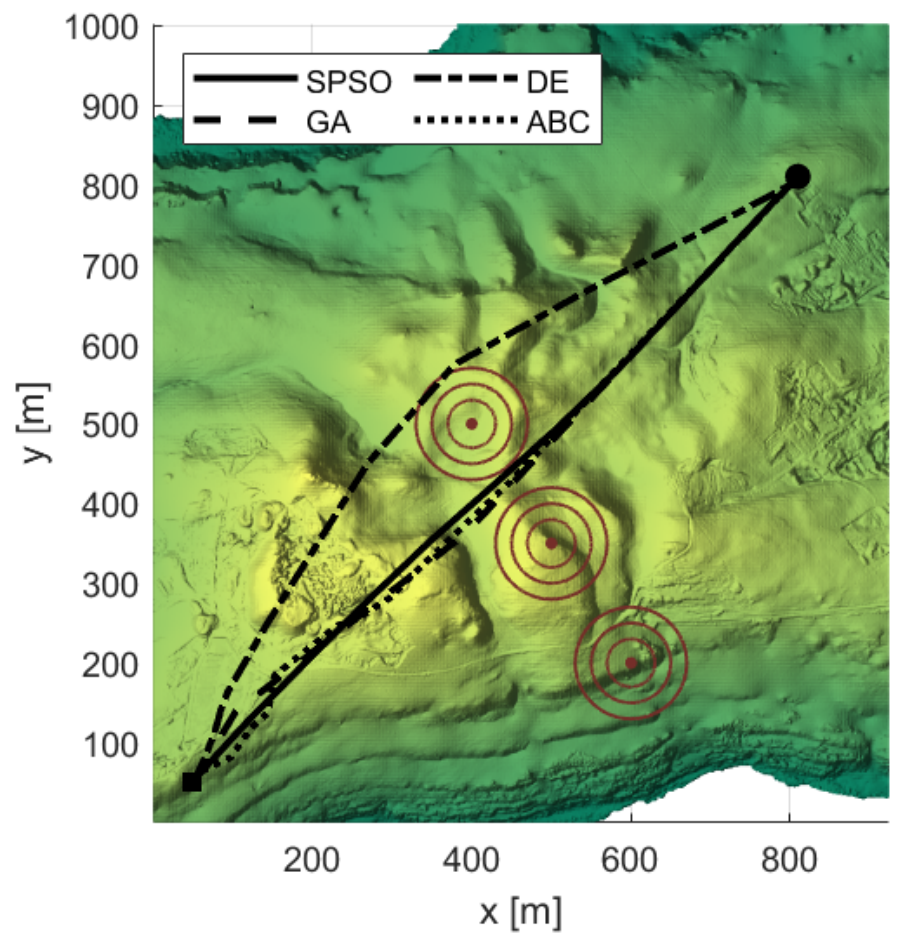}
		\caption{Scenario 6}
		\label{fig:toppath6b}
	\end{subfigure}
	\begin{subfigure}{0.5\textwidth}
		\centering
		\includegraphics[width=\textwidth]{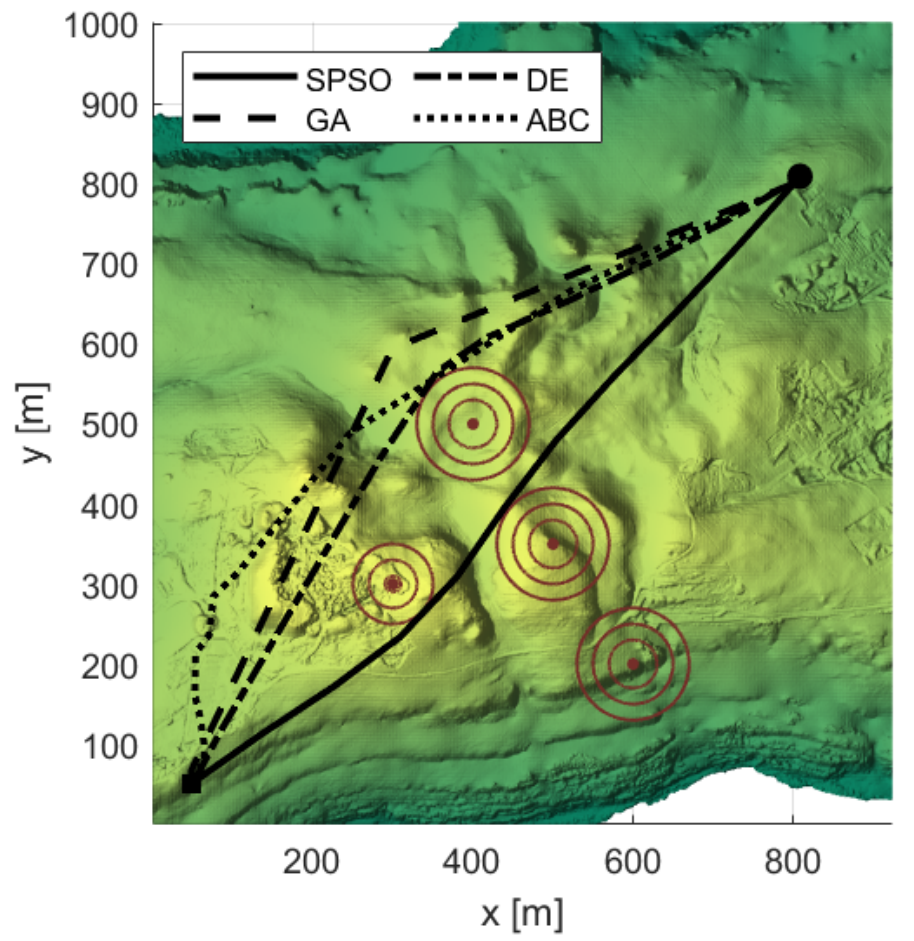}
		\caption{Scenario 7}
		\label{fig:toppath7b}
	\end{subfigure}%
	\begin{subfigure}{0.5\textwidth}
		\centering
		\includegraphics[width=\textwidth]{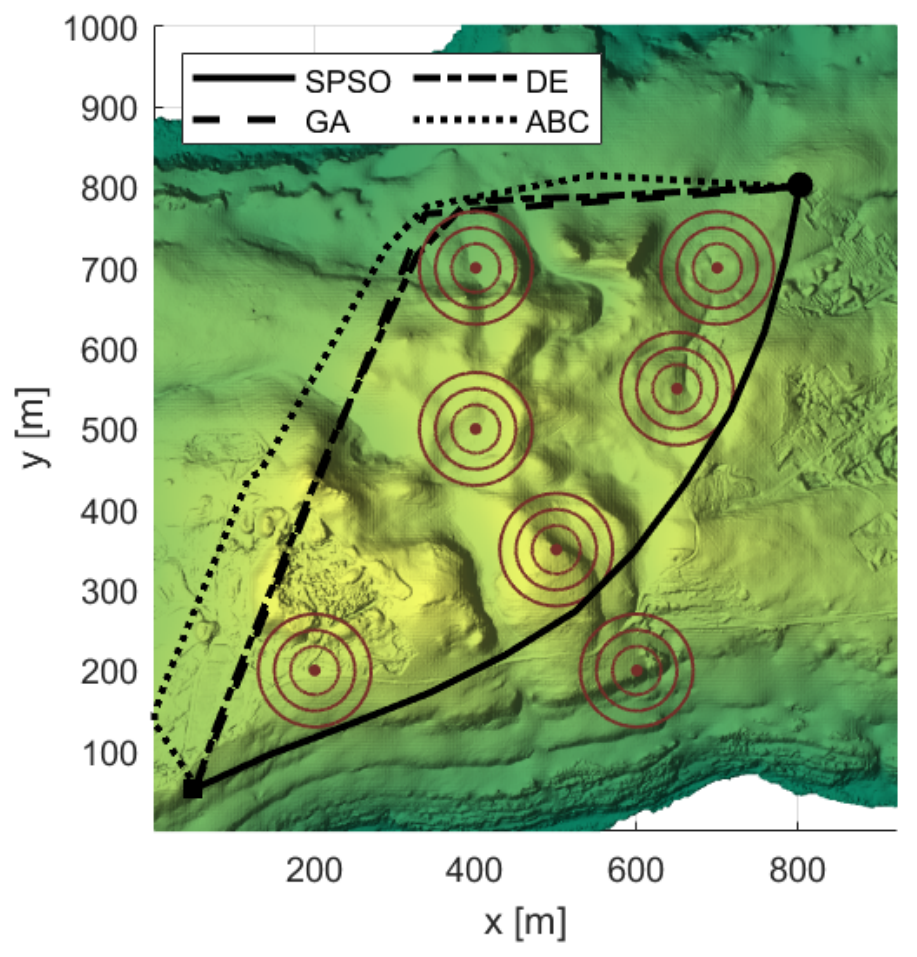}
		\caption{Scenario 8}
		\label{fig:toppath8b}
	\end{subfigure}
	
	\centering
	\caption{Top view of the paths generated by SPSO and other metaheuristic algorithms for scenarios 5 to 8}
	\label{fig:toppath58b}
\end{figure*}

Figures \ref{fig:toppath14b} and \ref{fig:toppath58b} provide the top view of the paths generated. It can be seen that all algorithms are able to generate collision free paths. However, SPSO introduces the smoothest and shortest paths in most scenarios. DE does not introduce the best paths for simple scenarios but provides near-optimal solutions for the complicated ones. It reflects the nature of DE that carries out exploration via mutation and selection to generate quality solutions but is limited in exploitation to find the optimal solutions. Similarly, ABC tends to generate average paths due to the compromise among different types of bees. Finally, GA generates paths that consist of only several line segments and sharp turns as the result of node deletion.

\begin{figure*}
	
	\begin{subfigure}{0.5\textwidth}
		\centering
		\includegraphics[width=\textwidth]{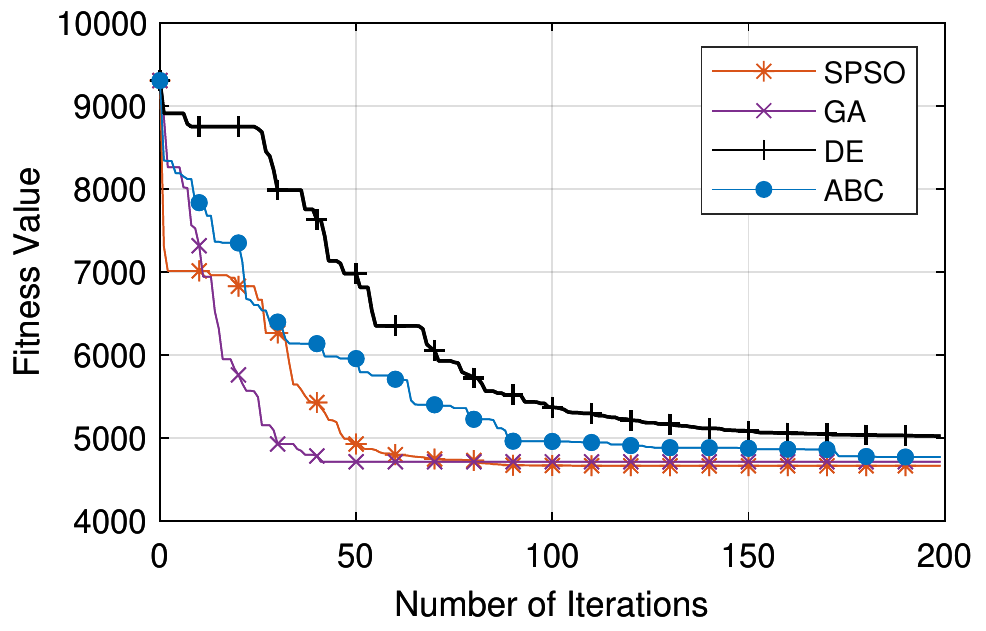}
		\caption{Scenario 1}
		\label{fig:best1b}
	\end{subfigure}%
	\begin{subfigure}{0.5\textwidth}
		\centering
		\includegraphics[width=\textwidth]{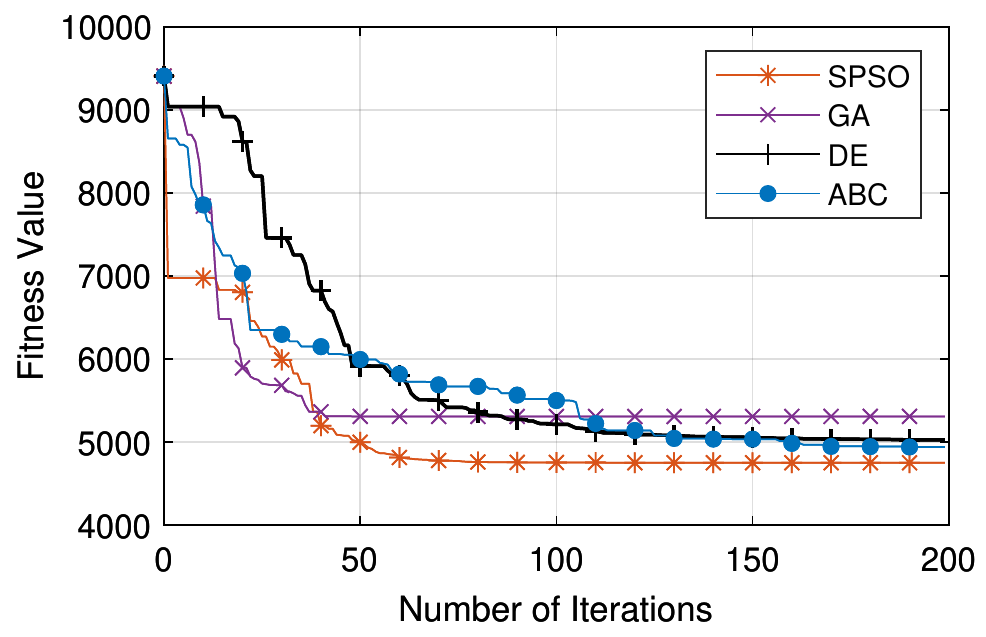}
		\caption{Scenario 2}
		\label{fig:best2b}
	\end{subfigure}
	\begin{subfigure}{0.5\textwidth}
		\centering
		\includegraphics[width=\textwidth]{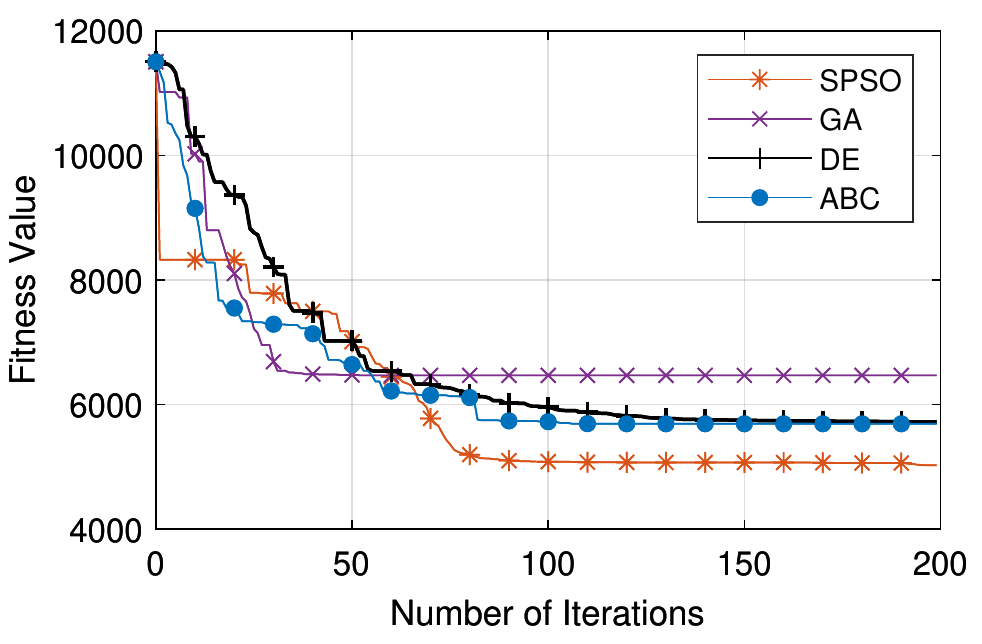}
		\caption{Scenario 3}
		\label{fig:best3b}
	\end{subfigure}%
	\begin{subfigure}{0.5\textwidth}
		\centering
		\includegraphics[width=\textwidth]{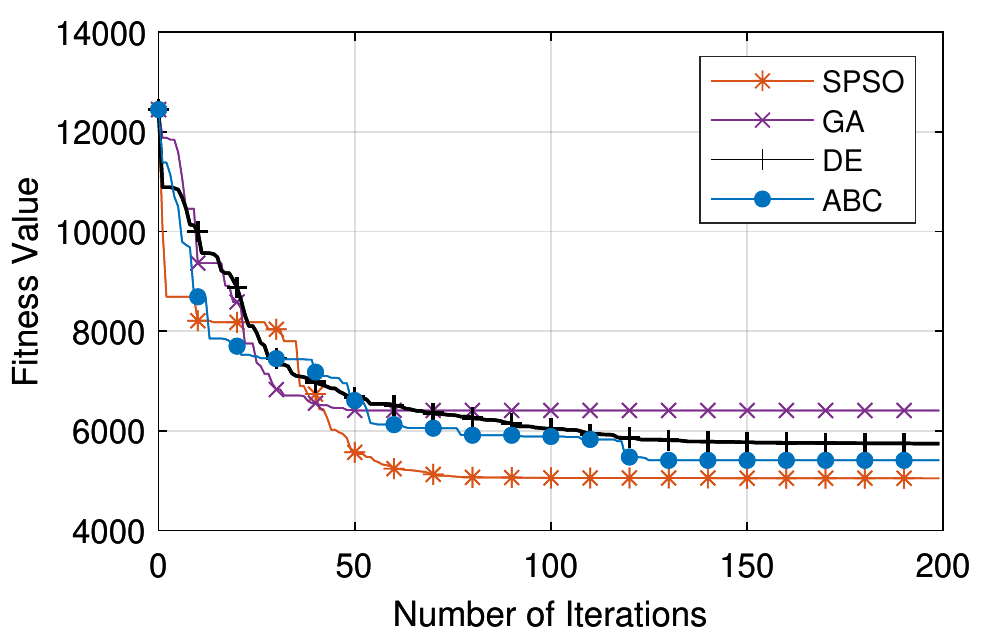}
		\caption{Scenario 4}
		\label{fig:best4b}
	\end{subfigure}
	\begin{subfigure}{0.5\textwidth}
		\centering
		\includegraphics[width=\textwidth]{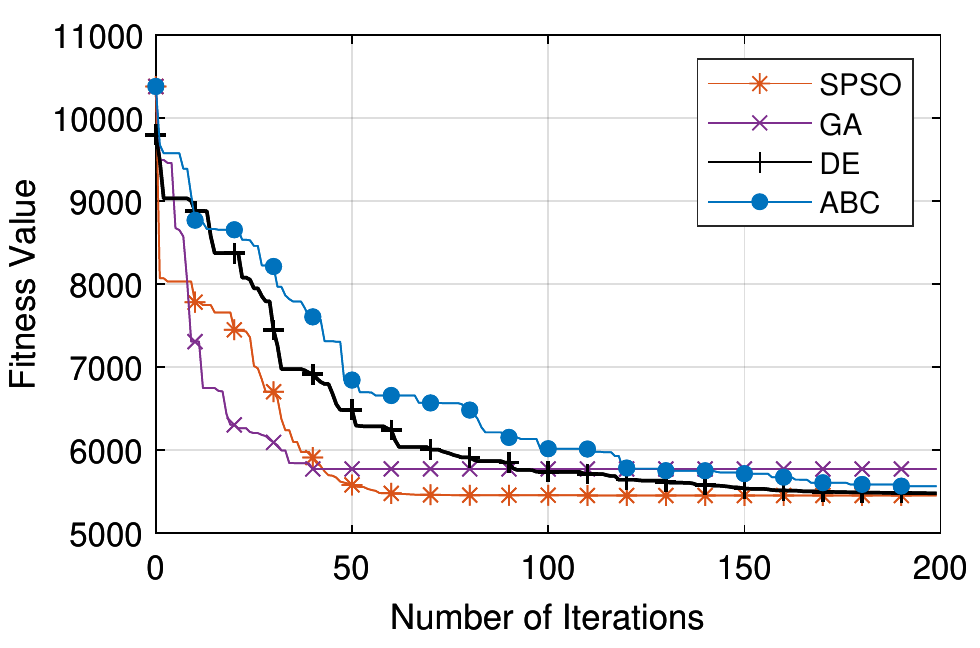}
		\caption{Scenario 5}
		\label{fig:best5b}
	\end{subfigure}
	\begin{subfigure}{0.5\textwidth}
		\centering
		\includegraphics[width=\textwidth]{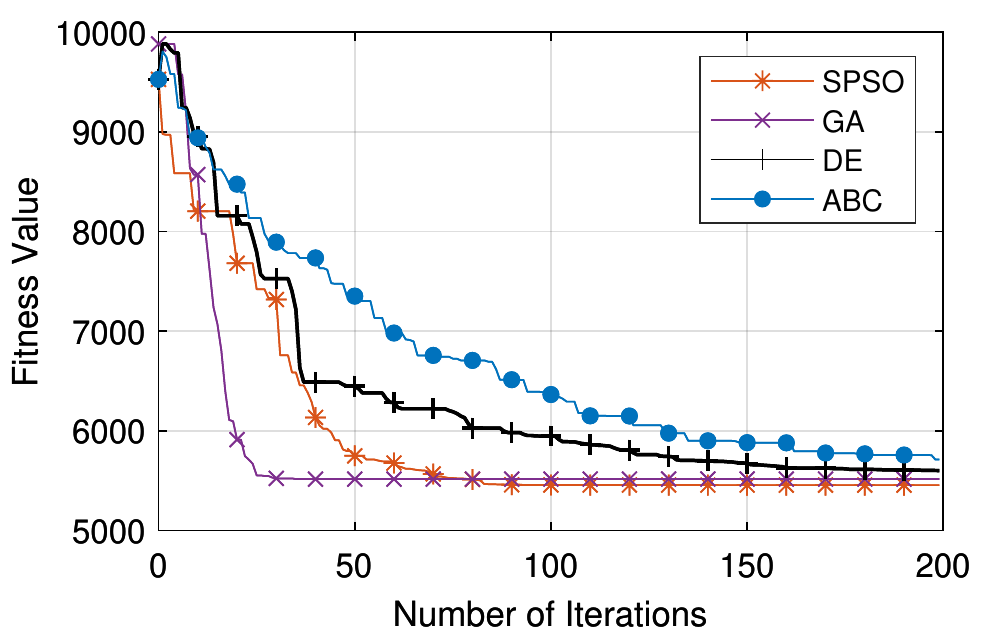}
		\caption{Scenario 6}
		\label{fig:best6b}
	\end{subfigure}		
	\begin{subfigure}{0.5\textwidth}
		\centering
		\includegraphics[width=\textwidth]{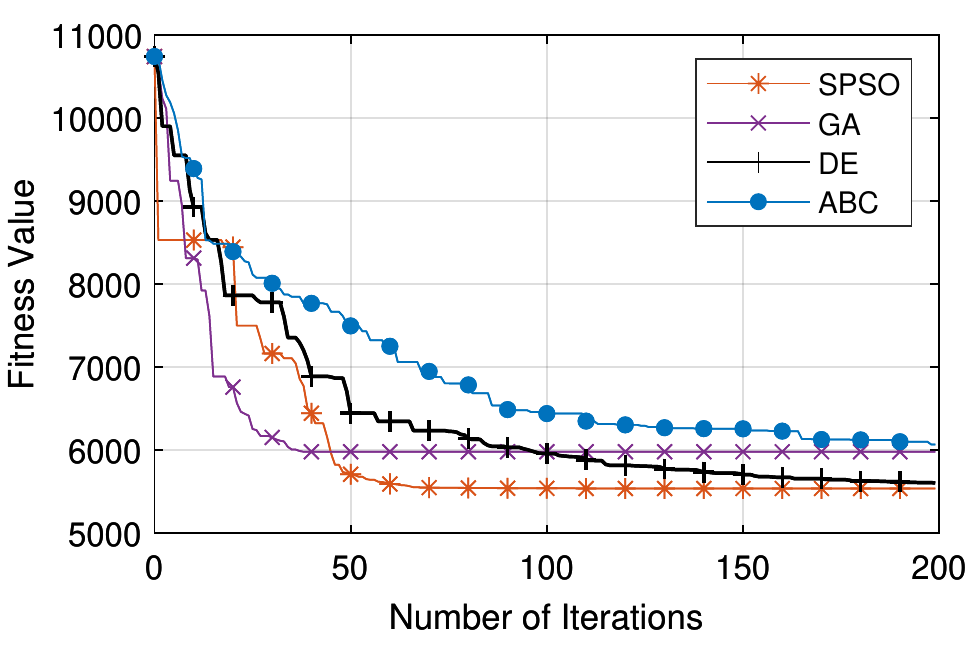}
		\caption{Scenario 7}
		\label{fig:best7b}
	\end{subfigure}
	\begin{subfigure}{0.5\textwidth}
		\centering
		\includegraphics[width=\textwidth]{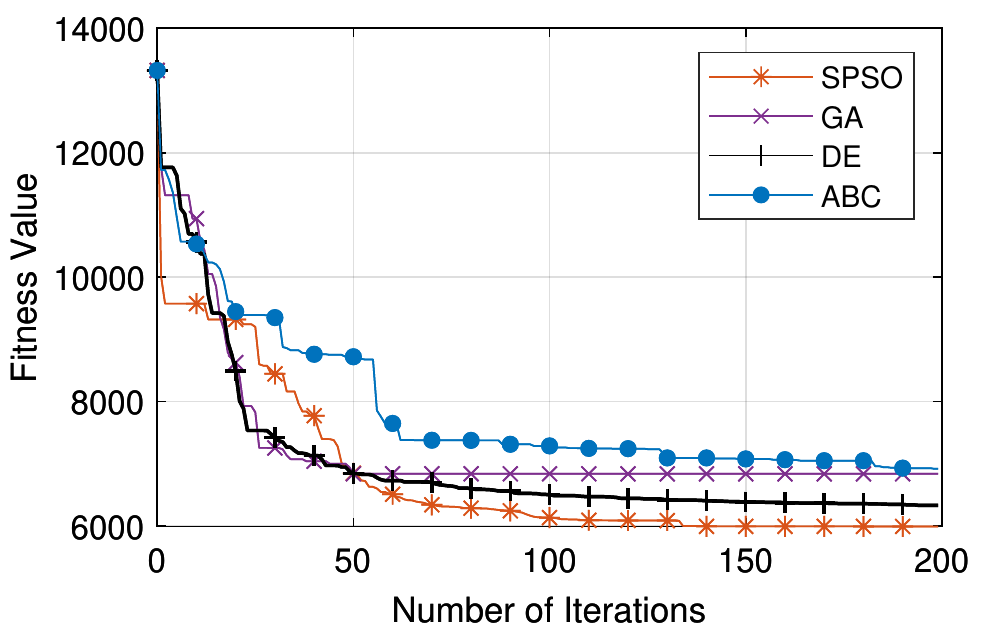}
		\caption{Scenario 8}
		\label{fig:best8b}
	\end{subfigure}	
	
	\centering
	\caption{Best fitness values over iterations of SPSO and other metaheuristic algorithms}
	\label{fig:bestCostb}
\end{figure*}

Figure \ref{fig:bestCostb} shows the best fitness over iterations where the values obtained by DE is scaled to 200 iterations for the sake of comparison. It can be seen that GA converges quickly to premature solutions due to search dimension reduction. ABC presents slow convergence because of its weakness in exploration. DE has steady convergence which shows the efficiency of its differential operator. Finally, SPSO has sufficiently fast convergence due to the balance between exploitation and exploration implemented via the social and coherence coefficients.

\subsection{Experimental verification}
\begin{figure*}	
	\begin{subfigure}{0.49\textwidth}
		\centering
		\includegraphics[width=0.9\textwidth]{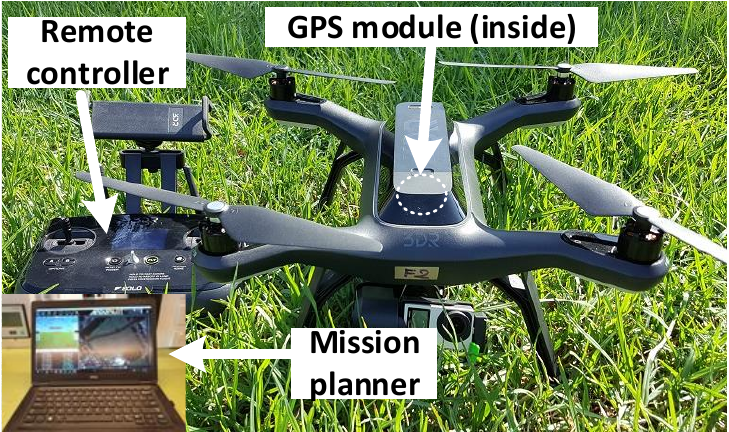}
		\caption{3DR Solo drone and Mission Planner software}
		\label{fig:drone}
	\end{subfigure}	%
	\begin{subfigure}{0.49\textwidth}
		\centering
		\includegraphics[width=1\textwidth]{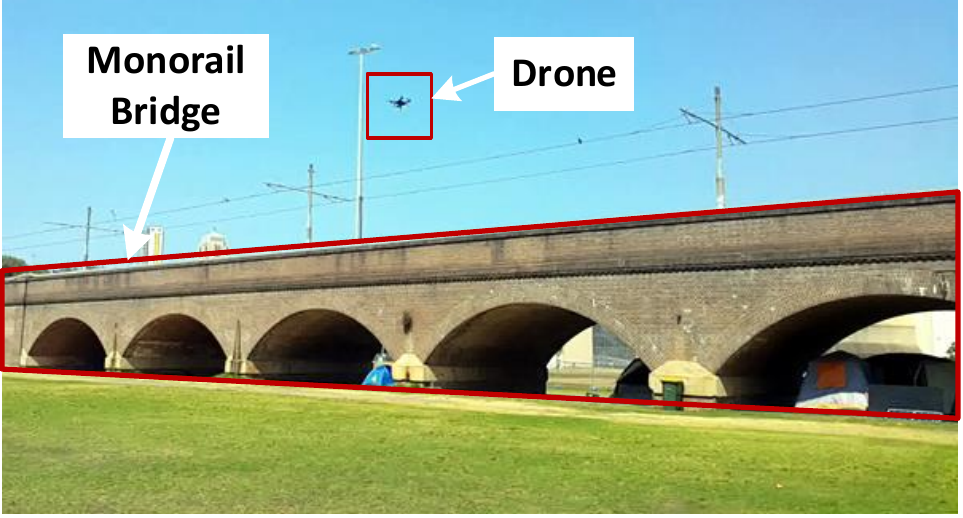}
		\caption{Experimental field with a monorail bridge}
		\label{fig:field}
	\end{subfigure}	
	\centering
	\caption{The drone and field used in experiments }
	\label{fig:experiment}
\end{figure*}

\begin{figure*}	
	\begin{subfigure}{0.5\textwidth}
		\centering
		\includegraphics[width=0.8\textwidth]{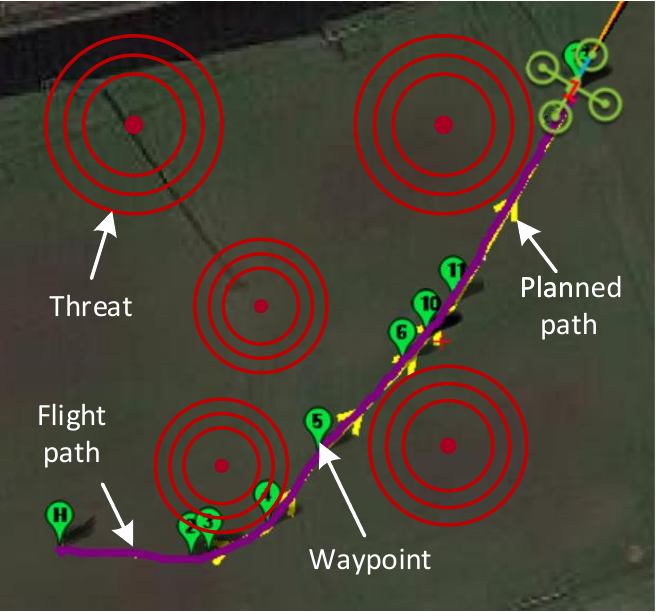}
		\caption{The planned (yellow) and actual flight (magenta) paths in experimental scenario 1}
		\label{fig:exp1}
	\end{subfigure}	
	\begin{subfigure}{0.5\textwidth}
		\centering
		\includegraphics[width=0.8\textwidth]{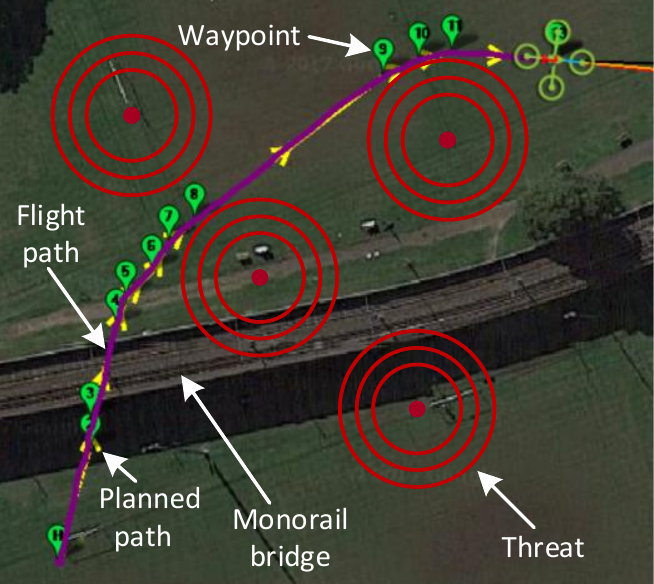}
		\caption{The planned (yellow) and actual flight (magenta) paths in experimental scenario 2}
		\label{fig:exp2}
	\end{subfigure}	
	\begin{subfigure}{0.5\textwidth}
		\centering
		\includegraphics[width=1\textwidth]{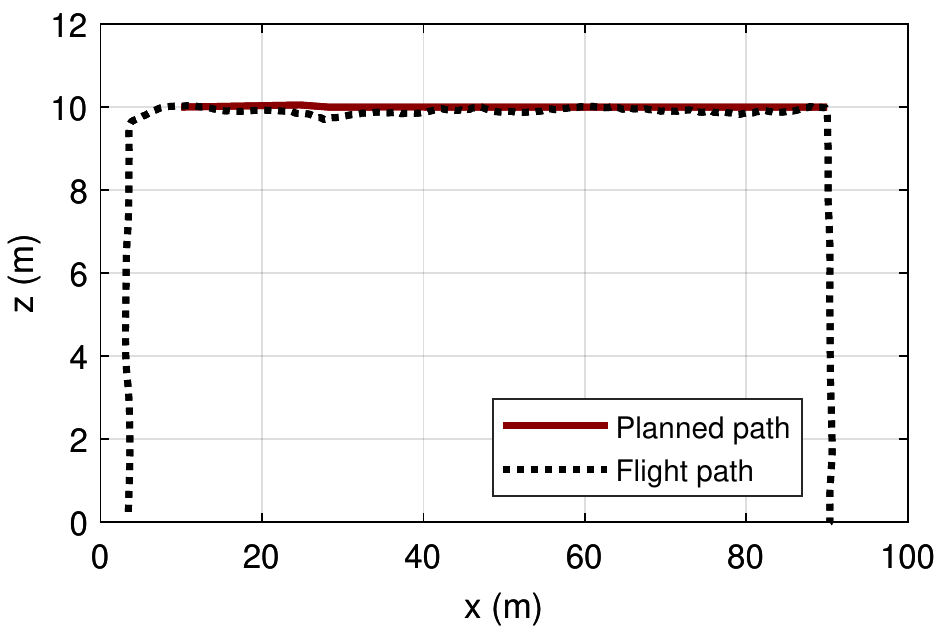}
		\caption{Altitude of the planned and actual flight paths in experimental scenario 1}
		\label{fig:exp1b}
	\end{subfigure}	
	\begin{subfigure}{0.5\textwidth}
		\centering
		\includegraphics[width=1\textwidth]{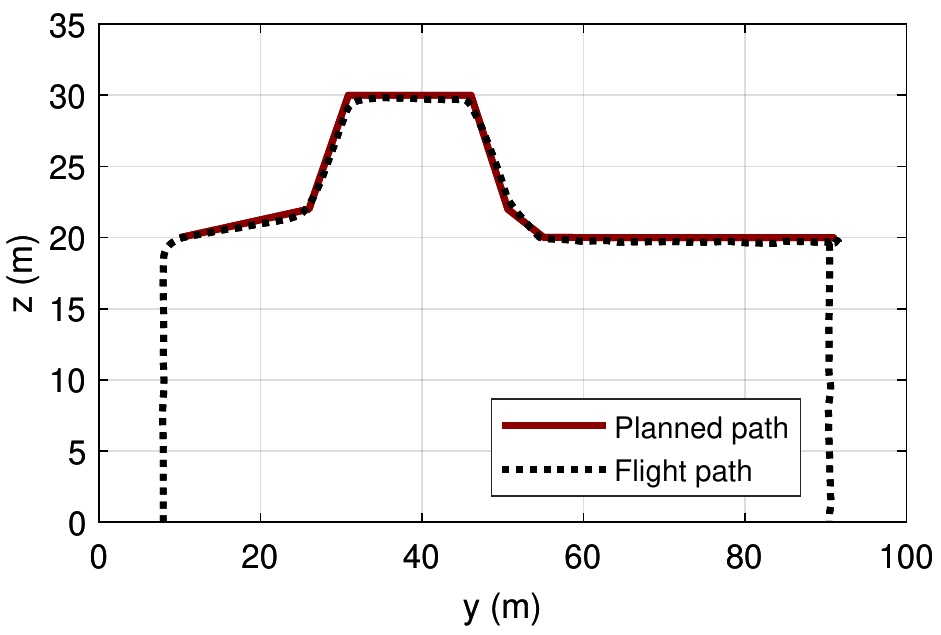}
		\caption{Altitude of the planned and actual flight paths in experimental scenario 2}
		\label{fig:exp2b}
	\end{subfigure}	
	
	\centering
	\caption{Experimental flight results }
	\label{fig:exp}
\end{figure*}

We have conducted several experiments to evaluate the validity of the generated paths for real UAV operations. The UAV used is a 3DR Solo drone that can be programmed to fly automatically via ground control station software named Mission Planner as shown in Fig. \ref{fig:drone}. The field used is a park in Sydney which has a monorail bridge that the drone needs to flight through as shown in Fig. \ref{fig:field}. The field is augmented with threats to create two experimental scenarios. Scenario 1 has a flat surface with 5 threats, whereas scenario 2 has 4 threats and includes the monorail bridge with sharp changes in height as shown in Fig. \ref{fig:exp1} and Fig. \ref{fig:exp2}. The longitude and latitude of the start and goal locations for scenario 1 is (-33.87643,151.191778) and (-33.875711,151.192643), and scenario 2 is (-33.875849, 151.191528) and (-33.87513,151.192394) respectively. Those locations, together with the terrain map and flight constraints are used as inputs of SPSO to generate waypoints. The waypoints are then uploaded to the drone via Mission Planner for autonomous flight.

Figure \ref{fig:exp1} shows the planned and flight paths obtained in real time from Mission Planner for scenario 1. It can be seen that the flight path is collision free and overlaps well with the planned path. The flight height, which is basically constant, also matches the planned path as shown in Fig. \ref{fig:exp1b}. Similar results have been obtained for scenario 2 as shown in Fig. \ref{fig:exp2} and Fig. \ref{fig:exp2b}. Notably, the drone can track the planned path to carry out abrupt changes in height to fly over the monorail bridge. Besides, the good match between the planned and flight paths indicates not only the validity of the path planning algorithm but also the accuracy of the positioning system implemented in the drone.   

\subsection{Discussion}

Through comparisons and experiments, it is clear that SPSO is capable of generating feasible, safe and optimal paths for UAV operation. The proposed algorithm performs especially well in complicated scenarios where many obstacles and threats appear reflected via its small fitness values and $D+$ $t$-test evaluations. The main drive for that effective performance rests with the idea of switching the search space, from the Cartesian to configuration space, where quality solutions can be obtained. Besides, constraints on UAV dynamics such as turning and climbing angles can be directly integrated into SPSO's variables to narrow down the search space. Nevertheless, hard constraints are being used in this study which may not be optimal for operations in which UAV states such as speed and altitude rapidly change over time. 

In our design, the cost function is scalable in the sense that additional requirements like fuel consumption can be added as a term $F_k$ with weight $b_k$ to the overall cost function (\ref{eq:cost}). However, choosing the right values of $b_k$ to reflect the relationship among requirements may become complicated as the number of requirements increases. In those situations, multi-objective optimization can be considered to fulfill the task. 

On another note, SPSO introduces relatively fast convergence as can be seen in Fig.\ref{fig:bestCost} and Fig.\ref{fig:bestCostb} due to coherent interactions among particles. However, like many other PSO variants, it faces the problem of premature convergence where its particles converge to a local optimum in certain scenarios. This is the case with scenario 7 where DE performs better than SPSO due to its exploitation capability obtained via mutation and recombination. It suggests that a relevant randomization mechanism such as mutation, random walk or L\'{e}vy flight \cite{HAKLI2014333} may be useful to deal with the premature convergence problem.

\section{Conclusion}
We have presented a new algorithm, SPSO, for the problem of UAV path planning with the focus on the safety and feasibility of the paths generated. The cost function is designed so that the constraints associated with optimality, safety and feasibility are simultaneously incorporated. SPSO is developed based on the correspondence between intrinsic motion components of the UAV and the search space. Comparisons on eight benchmarking scenarios generated from DEM maps show that SPSO achieves the best quality paths in most scenarios. PSO and $\theta$-PSO have stable convergence whereas QPSO only performs well for simple scenarios. Comparisons with other metaheuristic algorithms including GA, ABC, and DE also confirm the superior performance of SPSO. Experiments with real UAVs show the validity of the generated paths for practical operations. Besides, the correspondence between the particles of SPSO and UAV motion allows the kinematic constraints of UAV to be incorporated when necessary to further improve the path planning performance.

Our future work will focus on incorporating the exact constraints to be used for SPSO in the configuration space based on UAVs' dynamic model. We will also explore the applicability of SPSO to other optimization problems by evaluating its performance on different benchmarking functions.

%\section*{References}
\bibliography{reference}

\end{document}